\documentclass[lettersize,journal]{IEEEtran}
\usepackage{amsmath,amsfonts}
\usepackage{algorithm}
\usepackage{algpseudocode}
\usepackage{array}
\usepackage{textcomp}
\usepackage{stfloats}
\usepackage{url}
\usepackage{verbatim}
\usepackage{cite}
\usepackage{enumitem}
\usepackage{hyperref}
\usepackage{multirow}
\usepackage{comment}
\usepackage{xcolor}
\usepackage{float}
\usepackage{orcidlink}
\usepackage{booktabs}
\usepackage{graphicx} 
\usepackage[caption=false,font=footnotesize]{subfig}
\usepackage{xcolor}

\usepackage{xcolor}
\usepackage[normalem]{ulem}

\newcommand{\orcidID}[1]{\orcidlink{#1}}
\begin{document}

\title{Performance Drift Detection in Machine Learning as a Service (MLaaS) for IoT Environments}

\author{
Deepak~Kanneganti\,\orcidID{0009-0007-5860-2644},
Sajib~Mistry\,\orcidID{0000-0001-7513-3789},
Sheik~Mohammad~Mostakim~Fattah\,\orcidID{0000-0002-9103-6089},
Erik Elmroth\,\orcidID{0000-0002-2633-6798},
Aneesh~Krishna\,\orcidID{0000-0001-8637-5732},
and~Monowar~Bhuyan, \IEEEmembership{Senior Member,~IEEE}
\thanks{Deepak Kanneganti, Sajib Mistry, Sheik Mohammad Mostakim Fattah, and Aneesh Krishna are with the School of Electrical Engineering, Computing and Mathematical Sciences (EECMS), Curtin University, Perth, WA, Australia (e-mail: \{s.kanneganti, sajib.mistry, sheik.fattah, a.krishna\}@curtin.edu.au).}%
\thanks{Erik Elmroth and Monowar Bhuyan are with the Department of Computing Science, Ume{\aa} University, Ume{\aa}, Sweden (e-mail: \{elmroth, monowar\}@cs.umu.se).}
}

%\author{
%Deepak Kanneganti,
%Sajib Mistry,
%Sheik Mohammad Mostakim Fattah,
%Aneesh Krishna,
%and Monowar Bhuyan%
%\thanks{Deepak Kanneganti, Sajib Mistry, Sheik Mohammad Mostakim Fattah, and Aneesh Krishna are with the School of Electrical Engineering, Computing and Mathematical Sciences (EECMS), Curtin University, Perth, WA, Australia (e-mail: \{s.kanneganti, sajib.mistry, sheik.fattah, a.krishna\}@curtin.edu.au).}%
%\thanks{Monowar Bhuyan is with the Department of Computing Science, Ume\aa\ University, Ume\aa, Sweden (e-mail: monowar@cs.umu.se).}%
%}
\markboth{IEEE Trans. on Network and Service Management}%
{Kanneganti \MakeLowercase{\textit{et al.}}: Title of the Paper}


\maketitle

\begin{abstract}
Machine Learning as a Service (MLaaS) is a powerful cloud paradigm enabling data-driven intelligent applications in Internet of Things (IoT) environments, widely adopted across healthcare, smart homes, and industry due to its cost-effectiveness. However, the dynamic nature of IoT frequently alters data distributions, affecting MLaaS stability, while periodic MLaaS updates further introduce performance drift. Unlike traditional ML systems, MLaaS clients operate as black-box users without access to internal data or parameters, making drift detection particularly challenging. To address this, we propose a novel MLaaS Performance Drift Detection framework for IoT environments. The framework first employs an MLaaS extraction model that learns service behavior from input–output pairs and identifies prediction-influenced features. Building on this, the proposed MLaaS Performance Drift Detection (MPDD) model jointly captures variations in input data and MLaaS behavior. We further design an Adaptive-Temporal Performance Drift Detection Mechanism (APDDM) that dynamically adjusts monitoring frequency based on behavioral and data variations, enabling timely drift detection for effective service management. Extensive experiments on real-world datasets demonstrate that MPDD achieves up to 22–25\% accuracy improvement over baseline drift detection methods. APDDM provides an average accuracy gain of approximately 4\% and reduces the miss detection rate by around 9\% compared to fixed-interval monitoring. 

\end{abstract}

\begin{IEEEkeywords}
Machine Learning as a Service, IoT, Performance Drift, Drift Detection, Model Monitoring
\end{IEEEkeywords}

\maketitle
\section{Introduction}

\newcounter{sagemakerfn}

%These tools support workflows such as data preprocessing, model selection, hyperparameter tuning, and deployment.
 
Machine Learning as a Service (MLaaS) is a popular cloud-based service that provides scalable infrastructure and tools to build, train, and deploy ML models~\cite{ribeiro2015mlaas}. These services eliminate the complexity of managing hardware and manually designing ML models. MLaaS is typically accessed through web interfaces, application programming interfaces (APIs), or software development kits (SDKs). Major providers such as \textit{Microsoft Azure}\footnote{\url{https://azure.microsoft.com/en-us/products/machine-learning/}}, \textit{AWS SageMaker}\footnote{\url{https://aws.amazon.com/sagemaker/}}\setcounter{sagemakerfn}{\value{footnote}}, and \textit{OpenAI ChatGPT API}\footnote{\url{https://platform.openai.com/docs/}}
 offer comprehensive MLaaS solutions. MLaaS services are generally categorized as \textit{platform-based MLaaS (PMLaaS)} for centralized training and storage (e.g., \textit{AWS SageMaker}\footnotemark[\value{sagemakerfn}]), and \textit{inference-based MLaaS (IMLaaS)} for local prediction using pretrained models (e.g., \textit{Google Activity Recognition}\footnote{\url{https://developers.google.com/location-context/activity-recognition}}).

%MLaaS is widely adopted across domains such as the IoT, cybersecurity, and industrial automation. 

Internet of Things (IoT) companies often rely on IMLaaS to add intelligent capabilities without training models in-house~\cite{philipp2020machine}. For example, \textit{Medtronic}, a healthcare IoT provider, integrates IMLaaS from \textit{IBM IQ cast} into its application to predict low blood sugar events~\cite{medtronic2019sugariq}. Clients typically evaluate these services using ground truth data from their environments to assess functional attributes (e.g., model and data specifications) and quality-of-service metrics (e.g., efficiency and scalability). However, the performance of an IMLaaS model is not static and may change over time. This occurs mainly due to two factors: \textit{1) the dynamic nature of IoT environments}, where data patterns evolve~\cite{fadlullah2018delay}, and \textit{2) MLaaS updates}, where providers periodically update services beyond client control~\cite{medtronic2019sugariq}. As a result, clients may experience \textit{performance drift} over time.

\textit{Performance drift} refers to the gradual degradation of the predictive ability of IMLaaS over time, reflected in metrics such as accuracy, precision, or F1 score~\cite{baena2006early}. It occurs when the underlying relationship between input data \(X\) and output labels \(y\) changes, a phenomenon known as \textit{concept drift}~\cite{baena2006early,frias2014online}. IoT environments are highly dynamic due to evolving \textit{user behavior, sensor distributions, and system requirements}, making it difficult for clients to maintain the reliability of MLaaS services. For example, an IoT environment using Google’s Activity Recognition API for healthcare monitoring may face reduced accuracy as \textit{patient age, health condition, or mobility} alter sensor data over time, leading to performance drift. Additionally, MLaaS providers periodically update or retrain their models, which may further contribute to performance drift. Hence, IoT environments need to collect ground truth periodically to monitor MLaaS performance. However, collecting ground truth data requires substantial \textit{time and effort}, making it costly and often impractical. The key challenge is the \textit{timely detection of performance drift} without relying on continuous ground truth collection. To address this, we propose an \textit{MLaaS performance drift detection model} that operates without requiring continuous ground truth data.

Traditional ML drift detection methods typically rely on monitoring error-related measures (e.g., accuracy)~\cite{baena2006early,yan2020accurate,frias2014online,barros2017rddm}. These approaches assume timely access to ground-truth labels, which is often not feasible in real-world deployments. Consequently, label-free drift detection techniques focus on monitoring internal model parameters, training data characteristics, or input data distributions to identify shifts. For example, methods such as DRIFTLENS~\cite{greco2025unsupervised} and Type-LDD~\cite{yu2023type} analyze training data properties and internal model parameters (e.g., weights and gradients) to infer distributional changes, while other approaches rely solely on observable input data and use statistical tests or discriminative models to detect shifts in input data distributions~\cite{gozuaccik2019unsupervised,jang2022sequential,cerqueira2023studd}. However, these techniques may not be directly applied in MLaaS settings due to the black-box nature of MLaaS, where service providers do not expose internal parameters or training data because of security and proprietary constraints. Moreover, approaches that rely only on changes in input data characteristics mainly capture distributional variation and cannot determine whether such changes actually alter the predictive behavior of the deployed MLaaS service under evolving IoT streams. As a result, the relationship between changing IoT data and MLaaS output behavior remains unmodeled. Therefore, IoT environments require an effective mechanism that explicitly captures the interaction between input data evolution and observable MLaaS outputs to accurately identify performance drift and support reliable service management. To the best of our knowledge, performance drift detection in MLaaS remains an underexplored problem. We identify \textit{two} key challenges in detecting performance drift in MLaaS for IoT environments:

\begin{itemize}[itemsep=0ex, leftmargin=2ex, label=\textbullet]
    \item \textit{How to detect the MLaaS performance drift without ground truth?:} 
    The dynamic nature of IoT environments can cause changes in input data patterns, potentially affecting MLaaS performance~\cite{bayram2022concept,kanneganti2025adaptive}. IoT environments require continuous collection of ground-truth data to evaluate these services, which is time-consuming and often impractical. Existing studies often analyze statistical properties of input data distributions to detect drifts but distributional changes do not necessarily imply a performance change in MLaaS~\cite{greco2025unsupervised}. We consider this phenomenon as \textit{pseudo drift} (\textit{see Definition~3}). In addition, MLaaS providers periodically update or retrain their models, whereas IoT environments lack access to internal training data, feature distributions, or parameter updates. This highlights the need for mechanisms that \textit{jointly capture input data changes and MLaaS behavior} to distinguish performance drift without relying on ground truth.
    \item \textit{How frequently should the MLaaS be re-evaluated?:} 
    The second critical challenge is determining the appropriate time period for monitoring MLaaS. Since MLaaS performance may degrade due to changes in input data distribution and MLaaS behavior, using fixed or inappropriate intervals can lead to \textit{oversensitivity (short intervals)} or \textit{under-sensitivity (long intervals)} in detecting performance drift. For example, if the MLaaS is evaluated every five minutes, the drift detection technique may continuously generate pseudo-drift notifications, reflecting an oversensitive setting that incurs unnecessary computational cost and resource consumption. Conversely, if evaluation occurs only every few hours, the system may fail to capture timely variations and detect real drift too late. This highlights the need for an \textit{adaptive time-variable mechanism} that can adjust its evaluation period to detect performance drift and support effective service management decisions in IoT environments.
\end{itemize}

To address this challenge, \textit{we propose an MLaaS performance drift detection framework} that identifies performance degradation without continuously collecting ground truth. The framework is built upon an \textit{MLaaS Performance Drift Detection Model (MPDD)}, which determines when the MLaaS should be re-evaluated to detect potential drift. First, we design an MLaaS Extraction Model that leverages IoT input data and MLaaS outputs to mimic MLaaS behavior and identify key input features without requiring access to black-box MLaaS parameters or training data. Next, we design two complementary scores, the \textit{Fréchet Data Drift Score (FDDS)} and the \textit{MLaaS-Aware Drift Exposure Score (MDES)}, to capture changes in input data and their impact on MLaaS behavior. We then propose the \textit{MLaaS Performance Drift Detection Model}, which captures the relationship between input data and MLaaS behavior, enabling detection of both real and pseudo drift without relying on ground-truth labels. Finally, we propose an \textit{Adaptive-Temporal Performance Drift Detection Mechanism (APDDM)} that dynamically adjusts the evaluation interval to mitigate oversensitivity and undersensitivity, supporting effective service management decisions in IoT environments. Our contributions are summarized as follows:

%To address this challenge, \textit{we propose an MLaaS performance drift detection framework} that effectively identifies performance degradation without relying on continuously collecting the ground truth. The framework is built upon an \textit{MLaaS Performance Drift Detection Model (MPDD)}, which detects and determines when to re-evaluate the MLaaS to detect potential drift. First, we design an MLaaS Extraction Model that leverages IoT input data and MLaaS outputs to mimic MLaaS behavior, identifying key input features without requiring access to black-box MLaaS parameters or training data. Next, we design two complementary scores the \textit{Fréchet Data Drift Score (FDDS)} and the \textit{MLaaS-Aware Drift Exposure Score (MDES)} to capture changes in input data and their corresponding impact on MLaaS behavior. Next, we propose the \textit{MLaaS Performance Drift Detection Model}, which captures the relationship between input data and MLaaS behavior, enabling the detection of both real and pseudo drift without relying on ground-truth labels. Finally, we propose an \textit{Adaptive-Temporal Performance Drift Detection Mechanism (APDDM)} that dynamically adjusts the evaluation interval to mitigate oversensitivity and undersensitivity, supporting effective service management decisions in IoT environments. Our contributions can be highlighted as follows:
\begin{itemize}[itemsep=0ex, leftmargin=2ex, label=\textbullet]
\item We propose an MPDD framework designed for IoT environments that identifies performance degradation in black-box MLaaS models without relying on ground truth data.
\item We develop an MLaaS Extraction Model to approximate MLaaS behavior using input–output pairs.
%\item We introduce two drift measures — Fréchet Data Drift Score (FDDS) and MLaaS-Aware Drift Exposure Score (MDES); FDDS quantifies input data changes, while MDES combines feature importance and data drift to distinguish real from pseudo drift.
\item We propose a novel MPDD model  that jointly captures the relationship between input data and MLaaS behavior, allowing accurate identification of performance drift.
\item We propose an APDDM that dynamically adjusts evaluation intervals to balance oversensitivity and undersensitivity.
\end{itemize}

\vspace{-2mm}
\section{Prior Work}
%\mb{These two categories don't connect to the claims and results. Further, I suggest keeping the main paper within 12 pages. Some text can move to the Appendix, which is fine.}
\subsection{Error-Rate-Based Drift Detection}
ML services deployed in IoT environments often suffer from performance degradation over time due to changes in the underlying data-generating process, commonly referred to as concept drift \cite{tsymbal2004problem}. Prior research in IoT data stream learning has proposed several drift detection techniques that mainly rely on observable error-rate signals \cite{baena2006early,yan2020accurate,fattah2020event}. Classical error-monitoring detectors, such as Drift Detection Method (DDM) and Early Drift Detection Method (EDDM) \cite{baena2006early}, along with adaptive extensions including Reactive Drift Detection Method (RDDM) and Adaptive Windowing (ADWIN) \cite{barros2017rddm,bifet2007learning}, identify drift by monitoring variations in prediction errors within sliding or adaptive windows. More advanced variants, including Hoeffding-based approaches such as HDDM and FHDDM \cite{frias2014online,pesaranghader2016fast}, improve sensitivity through statistical testing based on Hoeffding bounds \cite{yan2020accurate}. Recent studies have also explored drift-aware analytics in IoT streaming environments. Yang and Shami~\cite{yang2021lightweight} proposed a lightweight IoT anomaly detection framework using sliding-window monitoring and dynamic retraining for evolving sensor and network traffic streams. Similarly, FedConD~\cite{chen2021asynchronous} introduced a federated IoT sensor learning framework that detects drift by monitoring historical prediction behavior from distributed local models. Despite addressing IoT-oriented streaming scenarios, these approaches still depend on observable prediction errors and continuous access to ground-truth labels \cite{baena2006early}. Such assumptions are often impractical in real-world IoT deployments due to the cost, delay, and difficulty of obtaining reliable labeled feedback from distributed sensor devices \cite{shayesteh2022automated,chiang2024detection}.

\subsection{Black-Box, Label-Independent Drift Detection Methods}
Input data and distribution-based detection methods focus on analyzing changes in the input data stream rather than output errors. These techniques rely on statistical divergence measures such as Maximum Mean Discrepancy (MMD), Kolmogorov–Smirnov~\cite{massey1951kolmogorov}, and Fréchet (Wasserstein-2) distance~\cite{vaserstein1969markov} to identify distributional shifts. MMD-based approaches compare data segments over time but often require large observation windows, limiting responsiveness in black-box environments~\cite{sinn2012detecting}. Similarly, DRIFTLENS~\cite{greco2025unsupervised} detects drift using training data representations and internal model parameters for unsupervised latent feature analysis, yet its effectiveness depends on access to model internals and training information, which is typically unavailable in service-oriented settings. The Type-LDD framework~\cite{yu2023type} formulates drift detection using a multi-task sharing-loss function pre-trained on synthetic data to capture drift timing and type, but assumes availability of training data and semantic knowledge aligned with the model structure.

Recent studies have also explored label-independent drift handling in IoT and edge environments under limited observability. For example,~\cite{pashamokhtari2023dynamic} detected concept drift by analyzing IoT traffic flow behavior and classification score distributions without requiring true labels, while~\cite{agate2024enhancing} combined KSWIN-based drift detection with unsupervised anomaly monitoring over IoT traffic streams. Similarly, A-Detection~\cite{wang2021concept} identified drift using reliability data streams collected from edge and IoT services. However, these approaches assume access to historical traffic distributions or reference data that are typically unavailable to clients. Existing black-box drift detection methods mainly focus on input-stream distributional changes rather than behavioral performance variations of deployed services. D3~\cite{gozuaccik2019unsupervised} detects drift by training a discriminative classifier to separate historical and recent input samples, while SCSD~\cite{jang2022sequential} models drift detection as a sequential classifier two-sample test to identify covariate shift through calibrated confidence intervals between training and live inputs. The STUDD framework~\cite{cerqueira2023studd} adopts a student–teacher paradigm, where drift is inferred by monitoring imitation loss as the student replicates teacher predictions under evolving inputs. Although these methods operate under limited observability, they primarily capture input-space variations rather than behavioral performance changes of deployed ML services. In MLaaS environments, clients only observe input–output interactions without access to training data or internal model states~\cite{chiang2024detection,barros2017rddm,bifet2007learning}. This limitation can lead to pseudo drift, where input distribution changes do not necessarily indicate actual service degradation. This gap motivates the need for methods that directly relate observable inputs to MLaaS behavioral performance, which we address through a \textit{performance drift detection framework} designed for black-box MLaaS settings.

\newcommand{\del}[1]{\textcolor{red}{\sout{#1}}}
\newcommand{\add}[1]{\textcolor{blue}{#1}}

\section{Motivation Scenario}
\begin{figure}[t]
    \centering
    \includegraphics[width=\columnwidth]{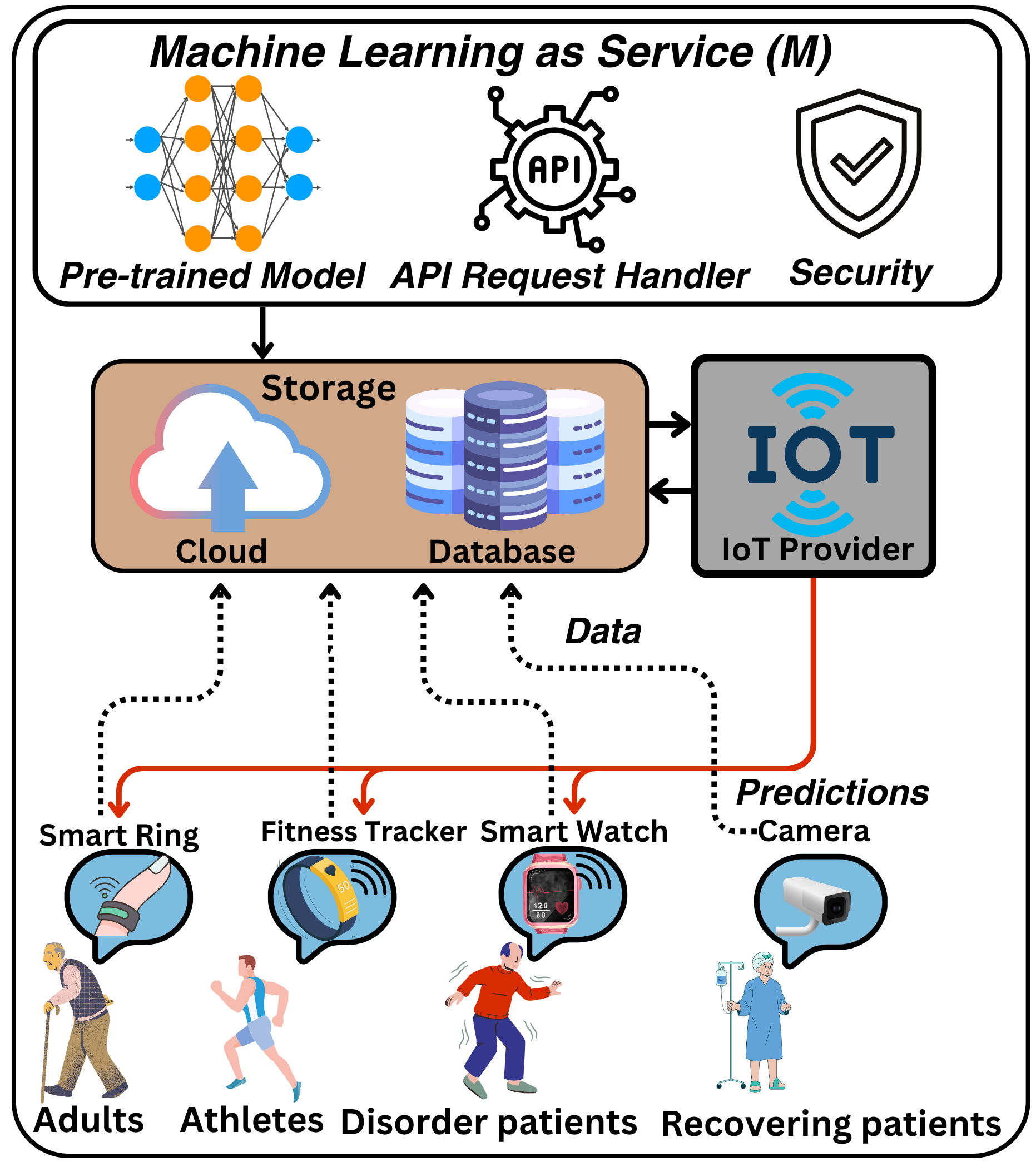}
    \caption{\normalsize Illustration of the Motivation Scenario}
    \label{fig2}
    \vspace{-5mm}
\end{figure}

Let us consider an IoT healthcare environment that aims to integrate Human Activity Recognition (HAR) capabilities to monitor and analyze patient activities such as walking, sitting, and sleeping. However, the environment lacks the expertise and infrastructure required to develop, train, and maintain an HAR model. Therefore, the environment leverages an inference-based MLaaS provider to obtain HAR predictions. Major cloud platforms such as Google Cloud offer pre-trained activity recognition models that can be deployed without in-house ML expertise. In practice, the IoT environment submits collected sensor data to the provider in periodic batches through batch inference channels such as \textit{Google Cloud Vertex AI Batch Prediction}\footnote{\url{https://cloud.google.com/vertex-ai/docs/predictions/batch-predictions}} or \textit{Azure Machine Learning batch endpoints}\footnote{\url{https://learn.microsoft.com/en-us/azure/machine-learning/concept-endpoints-batch}}. Fig.~\ref{fig2} illustrates the motivating scenario of MLaaS integration within the IoT.

% Let us consider an IoT healthcare environment that aims to integrate HAR capabilities to monitor and analyze patient activities such as walking, sitting, and sleeping. However, the environment lacks the expertise and infrastructure required to develop, train, and maintain an HAR model. Therefore, the environment leverages the capabilities of an inference-based MLaaS provider to obtain a pre-trained HAR model. For example, the STM32 AI Model Zoo provides pre-trained HAR models that can be deployed directly on edge for local predictions\cite{stm32_model_zoo}. Fig.~\ref{fig2} illustrates the motivating scenario of MLaaS integration within the IoT environment

Let us consider that the IoT environment is subject to changes influenced by user behavior, device configurations, and environmental factors. As a result, the current IoT environment may no longer reflect the conditions assumed during the original deployment of the MLaaS service. This phenomenon may lead to a noticeable degradation in performance reflected in metrics such as accuracy, precision, and recall. For instance, in healthcare, variations in patient demographics (e.g., pediatric to elderly), health conditions (e.g., recovery stages or mobility impairments), or sensor placements (e.g., wrist versus ankle) can alter sensor data and reduce prediction accuracy from 93\% to 68\%. Hence, the IoT environment requires a performance drift detection strategy for effective service management, enabling informed service-level decision making such as notifying the MLaaS provider or switching to a more suitable service.

The primary challenge is to detect performance drift in MLaaS without the ground truth. IoT environments are dynamic and subject to changes influenced by user behavior, device configurations, and environmental factors. However, a change in input data distribution does not necessarily indicate performance drift, as some MLaaS models may be robust enough to handle such variations, leading to pseudo drift. Therefore, detecting input changes alone does not solve the problem of identifying real performance drift. Moreover, MLaaS providers’ behavior must also be considered. Due to security restrictions, clients lack access to internal parameters or training data, making MLaaS a black-box model. This highlights the need for an effective mechanism to capture the underlying relationship between input data and MLaaS to detect real and pseudo drift without relying on ground truth.

Let us assume that an IoT environment designs a performance drift detection model to monitor MLaaS. Identifying an appropriate time interval for re-evaluation is a key challenge, as inappropriate choices can lead to \textit{oversensitivity (short intervals)} or \textit{undersensitivity (long intervals)} in detecting performance drift. For example, frequent checks (e.g., every minute) may trigger repeated pseudo-drift notifications, while infrequent checks (e.g., hourly) may delay the detection of real drift, allowing true performance degradation to persist unnoticed. Such inappropriate monitoring frequencies also increase computational and operational costs. Hence, IoT environments require a time-variable mechanism that can adaptively adjust the evaluation interval to detect real performance drift accurately and promptly.

\section{Key Definitions and Problem Statement}\label{MLaaS_problem}
\noindent\textbf{Definition 1: MLaaS.}
An MLaaS service is represented as a tuple $\langle F,\mathrm{QoS}\rangle$, where:
\begin{itemize}[itemsep=0ex, leftmargin=2ex, label=\textbullet]
    \item $F$ is the functional specification offered by $M$, including model specifications $\mathcal{N}_{\Phi}$ (e.g., model type $M_t$) and data specifications $D_s$, where $D_s=\{D_V,D_F\}$ denotes data volume ($D_V$), and data features ($D_F$).
    \item $\mathrm{QoS}$ is the non-functional specification offered by $M$, encompassing evaluation metrics ($E_f$, e.g., accuracy, $R^2$, task-specific scores), quality factor ($Q$, historical performance trends), and response time ($R_t$, latency).
\end{itemize}

\noindent\textbf{Definition 2: Real Drift.}
Consider an MLaaS service $M$. The stream of input--prediction pairs observed over the interval $[0,t]$ is defined as
\begin{equation}
S_t=\{(X_0,\hat{Y}_0),(X_1,\hat{Y}_1),\ldots,(X_t,\hat{Y}_t)\},
\end{equation}
where $X_i$ denotes the input instance at time $i$, and $\hat{Y}_i=M(X_i)$ represents the prediction generated by the MLaaS. Let $P_t^{M}(X,\hat{Y})$ denote the joint probability distribution of the input data and the corresponding MLaaS predictions over the interval $[0,t]$. A \textit{real drift} is said to occur at time $t+1$ if
\begin{equation}
P_t^{M}(X,\hat{Y}) \neq P_{t+1}^{M}(X,\hat{Y}),
\end{equation}
indicating a fundamental change in the relationship between the input data and the MLaaS predictions, thereby reflecting a degradation in predictive behaviour.

%\noindent\textbf{Definition 2. (Real Drift).}
%Consider a MLaaS service $M$ produce a sequence of predictions based on streaming input data at a given a time interval $[0, t]$ is defines as: 
%\[S_t = \{(X_0, \hat{Y}_0), (X_1, \hat{Y}_1), \ldots, (X_t, \hat{Y}_t)\}, \quad \text{where } \hat{Y}_i = M(X_i).\]
%where, $X_i$ denotes the input instance at time $i$, and $\hat{Y}_i = M(X_i)$ represents the predictions of the MLaaS. Let $P_t^M(X, \hat{Y})$ denote the joint probability distribution of the input data and the predictions from $M$ over the time interval $[0, t]$. A \textit{real drift} is said to occur at time $t+1$ if: \[P_t^M(X, \hat{Y}) \neq P_{t+1}^M(X, \hat{Y}),\] 

%\( P_t^M(X, \hat{Y}) \) is the joint distribution of input data and model predictions over the interval $t$ reflecting a fundamental degradation in the MLaaS's predictive ability.

%. A \textit{performance drift} is said to occur at time $t+1$ if this distribution changes, reflecting a degradation in the MLaaS predictive ability.

\noindent\textbf{Definition 3 (Pseudo Drift).} A \emph{pseudo drift} occurs when the input data distribution changes while the underlying predictive mapping remains stable, such that MLaaS performance does not experience sustained degradation.
\begin{equation}
P_t^{M}(X) \neq P_{t+1}^{M}(X)
\quad \text{and} \quad
P_t^{M}(\hat{Y}\mid X) \approx P_{t+1}^{M}(\hat{Y}\mid X),
\end{equation}
where $P_t^{M}(X)$ denotes the input data distribution at time $t$, and $P_t^{M}(\hat{Y}\mid X)$ represents the conditional predictive distribution of the MLaaS. A pseudo drift reflects a temporary fluctuation in input data that does not alter the predictive mapping of the service and therefore does not lead to long-term performance degradation.

%Equivalently, a pseudo drift is a short-lived disturbance in the combined behavior of inputs and predictions, where the system temporarily deviates from its usual pattern but subsequently stabilises and returns to its original state without lasting impact on performance.

\noindent\textbf{MLaaS Performance Drift Detection Problem}.
The MPDD problem concerns automatically distinguishing between \textit{real drift} and \textit{pseudo drift} in an MLaaS service when changes in the operational environment are observed. To this end, the drift detection model outputs a drift classification (DC) and is defined as the following function
\begin{equation}
\mathrm{MPDD}(M,S_t)=
\begin{cases}
1, & \text{if } P_t^M(X,\hat{Y}) \neq P_{t+1}^M(X,\hat{Y}),\\
0, & \text{if } P_t^M(X) \neq P_{t+1}^M(X).
\end{cases}
\label{eq:drift_class}
\end{equation}
Here, $M$ denotes the MLaaS service in operation, and $S_t=\{(X_i,\hat{Y}_i)\}_{i=0}^{t}$ represents the stream of recent input–prediction pairs up to time $t$. A classification of \textit{Real Drift} requires immediate intervention (e.g., model retraining or service reconfiguration), whereas \textit{Pseudo Drift} indicates that the service can continue operating without corrective action.

\begin{table}[t]
\centering
\caption{Notation Summary}
\label{tab:notation_summary}
\renewcommand{\arraystretch}{1.1}
\setlength{\tabcolsep}{6pt}
\begin{tabular}{p{0.22\linewidth} p{0.70\linewidth}}
\hline
\textbf{Notation} & \textbf{Description} \\
\hline
$M$ & Black-box ML service (MLaaS model). \\
$F$ & Functional specification of the ML service. \\
$\mathrm{QoS}$ & Non-functional (quality) attributes of the service. \\
$D_s$ & Data specification of the service. \\
$X$ & Input feature vector. \\
$d$ & Number of input features. \\
$x_i$ & $i$-th input feature. \\
$b$ & Baseline (reference) window. \\
$w$ & Current observation window. \\
$I(x_i)$ & Feature importance score of $x_i$. \\
$I$ & Feature importance vector. \\
$\delta$ & Small perturbation applied to input features. \\
$\theta_m$ & Drift decision threshold. \\
$m_t$ & Drift score at block $t$. \\
$B$ & Buffer of recent drift scores. \\
$W_s$ & Adaptive window size. \\
$W_{\min}, W_{\max}$ & Minimum and maximum window bounds. \\
$\Delta W$ & Window adjustment step. \\
$L$ & Adaptation trend length. \\
$c_u$  & Counter for drift sensitivity (oversensitivity). \\
$c_o$ & Counter for non-drift sensitivity (undersensitivity). \\
$r$ & Drift ratio within the buffer. \\
$\theta_r$ & drift ratio threshold.\\
\hline
\end{tabular}
\end{table}

\section{Proposed MLaaS Performance Drift Detection Framework for IoT environments}

\begin{figure*}[ht]
    \centering
    \includegraphics[width=0.9\textwidth]{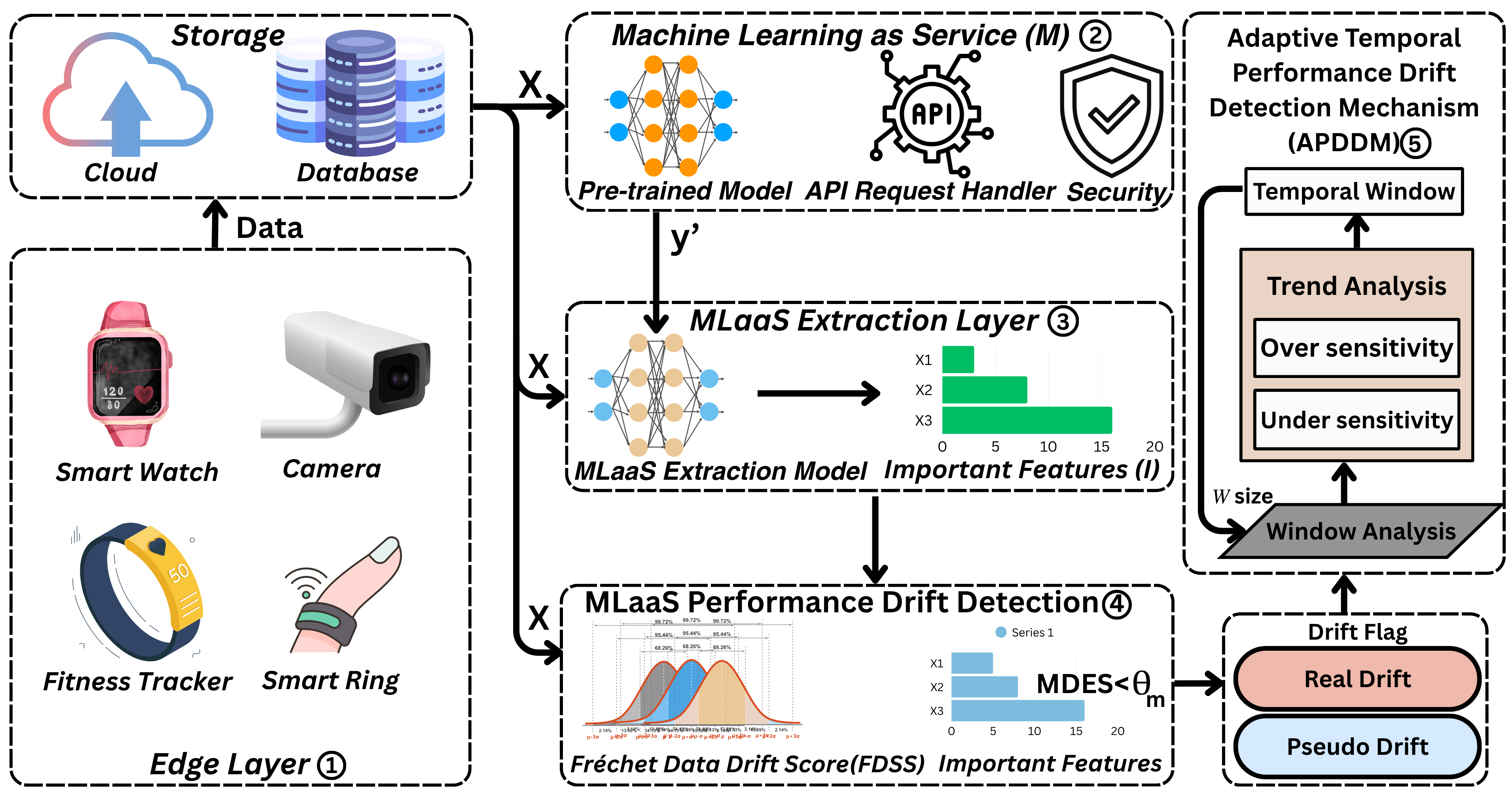}
    \caption{\normalsize Machine Learning as a Service (MLaaS) Performance Drift Detection Framework for IoT environments}
    \label{fig1}
\end{figure*}

%In this section, we present the MLaaS Performance Drift Detection (MPDD) framework, designed to identify performance drift in MLaaS within IoT environments. The framework comprises five key components, as illustrated in Fig.~\ref{fig1}. The first component is the edge and database layer, where IoT devices such as smartwatches, fitness trackers, cameras, and smart rings continuously generate real-time data streams stored in a centralized database. The second component, the MLaaS layer, integrates a pre-trained model provided by the MLaaS provider into the IoT edge to perform local predictions (\textit{see Definition 1}). The third component, the MLaaS extraction layer, trains an extraction model to capture latent factors influencing predictions. The performance drift detection module then compares the input data statistics with extraction model–derived features over time to differentiate between real and pseudo drift. Lastly, the adaptive temporal performance drift detection mechanism dynamically adjusts the time interval, ensuring reliable drift detection while reducing oversensitivity to minor fluctuations.

In this section, we present the MLaaS Performance Drift Detection (MPDD) framework for identifying performance drift in MLaaS within IoT environments. The framework comprises five key components, as illustrated in Fig.~\ref{fig1}. The first component is the edge and database layer, where IoT devices such as smartwatches, fitness trackers, cameras, and smart rings generate real-time data streams stored in a centralized database. The second component, the MLaaS layer, integrates a pre-trained model provided by the MLaaS provider into the IoT edge for local predictions (\textit{see Definition 1}).  The third component, the MLaaS extraction layer, trains an extraction model to capture latent factors influencing predictions. The extraction model $M'$ is constructed using queried MLaaS input--output interactions collected from the IoT environment and operates within the cloud infrastructure. In the current implementation, the extraction model is initialized through an offline training phase prior to runtime monitoring. The performance drift detection module compares input-data statistics with extraction models–derived features over time to distinguish real and pseudo drift. Finally, the adaptive temporal performance drift detection mechanism dynamically adjusts the monitoring interval, ensuring reliable drift detection while reducing oversensitivity to minor fluctuations.

\vspace{-3mm}
\subsection{MLaaS Extraction model} 
We propose an \textit{MLaaS Extraction Model} to approximate MLaaS behaviour using unlabeled input instances and the corresponding predictions returned by the service, as illustrated in \textit{Algorithm~\ref{alg:mlaas_extraction}}. We assume that the IoT provider maintains a historical repository of unlabeled inputs collected during normal system operation. Accordingly, Phase~1 (\textit{Input--Output Query Log Collection}) constructs the input--output query log from these historical inputs for offline training of the MLaaS Extraction Model. Once trained, the proposed framework requires no additional MLaaS queries beyond the application's normal inference requests. Traditional model-extraction approaches often rely on prediction probabilities, which can be affected by protection mechanisms such as output rounding, quantization, and probability perturbation \cite{tramer2016stealing,tang2024modelguard}.To mitigate the impact of these protection mechanisms, the proposed MLaaS Extraction Model relies exclusively on predicted class labels. The resulting input--label pairs capture the observable decision behaviour of the MLaaS and are used to train the extraction model. Since surrogate-based methods (e.g., \textit{SHAP}) incur high computational overhead in MLaaS settings, Phase~2 (\textit{Extraction Model Training}) trains a lightweight decision-tree model $M'$ using the collected query log to approximate the observable input--output behaviour (\textit{lines~8--9}). In our implementation, $M'$ is a \textit{CART decision-tree classifier} trained using the Gini impurity criterion and best-split strategy. The maximum tree depth is set to four, with a minimum split size of two samples and a minimum leaf size of one sample. The model is trained using MLaaS-predicted labels from the query log. To capture both stable and drifted operating conditions, the query pool consists of 60\% clean samples and 40\% drift samples, with drift data uniformly collected from sudden, gradual, incremental, and recurrent drift scenarios. The detailed configuration and training setup of $M'$ are summarized in Table~\ref{tab:extraction_configuration}. In Phase~3 (\textit{Feature-Importance Extraction}), the trained model perturbs each input feature individually and measures the output deviation to compute feature-importance scores as the average deviation across samples (\textit{lines~10--17}). Finally, the algorithm returns the aggregated feature-importance vector (\textit{line~18}), which forms the basis for subsequent performance drift analysis.

\textbf{Lemma 1}\label{lm1} \textit{Let $M$ be the black-box MLaaS service and $M'$ the extraction model. Assume that $M'$ satisfies a bounded fidelity condition such that \(
\mathbb{E}[|M(X) - M'(X)|] \leq \epsilon.
\) Then, the feature importance vector $\mathbf{I}^{M'}$ extracted from $M'$ consistently approximates the true behavioural importance $\mathbf{I}^{M}$ of $M$, and the deviation remains bounded as}
\[
\|\mathbf{I}^{M'} - \mathbf{I}^{M}\| \leq f(\epsilon, \delta),
\]
\textit{where $\epsilon$ denotes the fidelity error and $\delta$ denotes the perturbation sensitivity in the importance estimation process.}\\
\textit{Proof.} Please refer to \textit{Appendix A}.

\begin{algorithm}[t]
\caption{MLaaS Extraction Model}
\label{alg:mlaas_extraction}
\begin{algorithmic}[1]
\State \textbf{Input:} Unlabeled inputs $X=\{X^{(1)},\dots,X^{(N)}\}$, black-box MLaaS $M(\cdot)$
\State \textbf{Output:} Feature-importance vector $I=\{I(x_1),\dots,I(x_d)\}$
\Statex \textbf{Phase 1: Input--Output Query Log Collection}
\State Initialize $\mathcal{D} \gets \emptyset$
\For{$j \gets 1$ \textbf{to} $N$}
    \State $\hat{y}^{(j)} \gets M\!\left(X^{(j)}\right)$
    \State $\mathcal{D} \gets \mathcal{D} \cup \{(X^{(j)}, \hat{y}^{(j)})\}$
\EndFor
\Statex \textbf{Phase 2: Train Extraction Model $M'$}
\State Initialize $M'$ as a lightweight decision-tree model
\State Train $M'$ on $\mathcal{D}$ to approximate the mapping $X \rightarrow \hat{y}$
\Statex \textbf{Phase 3: Feature Importance Extraction}
\For{$i \gets 1$ \textbf{to} $d$}
    \State $I(x_i) \gets 0$
    \For{$j \gets 1$ \textbf{to} $N$}
        \State $X^{(j)}_{+i} \gets (x^{(j)}_1,\dots,x^{(j)}_i+\delta,\dots,x^{(j)}_d)$
        \State $I(x_i) \gets I(x_i) + \left\| M'(X^{(j)}_{+i}) - M'(X^{(j)})\right\|_1$
    \EndFor
    \State $I(x_i) \gets \frac{1}{N} I(x_i)$
\EndFor
\State \Return $I$
\end{algorithmic}
\end{algorithm}

\vspace{-3mm}
\subsection{Fréchet Data Drift Score (FDDS)}
\textit{Fréchet distance} (also known as the Wasserstein-2 distance), has been widely used to measure the distance between distributions of models’ features in deep learning studies \cite{dowson1982frechet}. In our context, we use it to measure the statistical difference between the historical (baseline) data used to select the MLaaS and the new data windows in the stream, and we call this metric the \textit{Fréchet Data Drift Score (FDDS)}. The proposed framework incorporates a preprocessing and feature transformation layer prior to FDDS computation. This layer transforms categorical and raw event-log data into structured numerical feature representations, enabling FDDS to operate on the transformed feature space. Given a multivariate normal baseline distribution $b$ characterized by mean vector $\mu_b$ and covariance matrix $\Sigma_b$, and a new data window $w$ with mean vector $\mu_w$ and covariance matrix $\Sigma_w$, the FDDS is computed as
\begin{equation}
\text{FDDS}(b_i,w_i)=
\underbrace{\lVert \mu_b-\mu_w \rVert_2^2}_{\text{Mean Shift}}
+
\underbrace{
\operatorname{Tr}\!\left(
\Sigma_b+\Sigma_w
-2\sqrt{\Sigma_b\Sigma_w}
\right)
}_{\text{Covariance Shift}},
\label{eq:fdds}
\end{equation}
where $\lVert \mu_b-\mu_w \rVert_2^2$ measures the squared difference between the mean vectors (shift in center), and the trace term measures the difference in the covariance structure (spread) between the two distributions. A higher FDDS indicates a larger divergence and a higher likelihood of data drift.\\
\textbf{Lemma 2}\label{l2}
\textit{Let $b$ and $w$ denote the baseline and the current data window as defined in the FDDS formulation in Eq.~(2). Since FDDS decomposes the statistical difference into mean shift and covariance shift, and the covariance shift term is non-negative, FDDS is lower bounded by the mean shift, i.e.,}
\[
\mathrm{FDDS}(b_i,w_i) \ge \|\mu_b - \mu_w\|_2^2.
\]
\textit{When the covariance structures are approximately unchanged, FDDS remains stable under small mean perturbations. Conversely, larger mean and/or covariance changes yield higher FDDS. Thus, FDDS stays low under minor deviations and increases under significant drift, ensuring effective drift differentiation between the baseline batch and the current window.}\\
\textit{Proof.} Please refer to \textit{Appendix A}.

\vspace{-3mm}
\subsection{MLaaS-Aware Drift Exposure Score (MDES)}
\textit{MDES} is designed to capture the relationship between changes in input data and MLaaS behavior. MDES quantifies the degree of exposure of an MLaaS service to performance drift by combining the \textit{MLaaS extraction feature importance} with the \textit{FDDS}. This integration allows the framework to capture changes in input data characteristics and variations in MLaaS behavior inferred through the MLaaS extraction model. Unlike traditional drift detectors that rely solely on distributional change, MDES considers how these changes interact with the influential features identified by the MLaaS extraction model, providing a deeper understanding of drift causes. Formally, let \( \mathbf{I} = \{ I(x_1), I(x_2), \ldots, I(x_d) \} \) represent the importance weights obtained from the MLaaS extraction model for \( d \) input features, and let \( \mathrm{FDDS}(b,w) \) denote the Fr\'{e}chet Data Drift Score between the baseline window \( b \) and the current window \( w \). The \textit{MDES} is defined as
\begin{equation}
\text{MDES}=
\begin{cases}
1, & \text{if } \left(\sum_{i=1}^{d} I(x_i)\,\operatorname{FDDS}(b_i,w_i)\right)>\theta_m, \\[2pt]
0, & \text{otherwise},
\end{cases}
\label{eq:MDES}
\end{equation}
Here, $\theta_m$ is an automated drift-exposure threshold derived from the mean ($\mu$) and standard deviation ($\sigma$) of the drift-exposure scores, defined as $\theta_m=\mu+\beta\sigma$. The sensitivity coefficient $\beta$ controls the threshold level and is empirically determined. When the weighted combination of feature importance and data drift exceeds $\theta_m$, the MLaaS service is classified as experiencing \textit{real drift}. Changes that do not exceed $\theta_m$ are classified as \textit{pseudo drift}

\textbf{Lemma 3}\label{l3}
\textit{Let $b$ be the baseline window and $w$ the current window, and let $\mathrm{FDDS}(b,w)$ denote the statistical shift between them. Let $\mathbf{I}=\{I(x_1),\ldots,I(x_d)\}$ be the bounded feature-importance vector obtained from the MLaaS extraction model, and let MDES be defined as in Eq.~(3). If the statistical shift captured by $\mathrm{FDDS}(b,w)$ is primarily concentrated on features with low importance weights, then the cumulative weighted exposure remains bounded and the drift is characterized as pseudo drift. Conversely, if the shift aligns with highly influential features, the weighted exposure increases proportionally and the drift is characterized as real drift.}
\textit{Proof.} Please refer to \textit{Appendix A}.

\begin{algorithm}[t]
\caption{MLaaS Performance Drift Detection Model}\label{alg:made}
\begin{algorithmic}[1]
\State \textbf{Input:} Input data $(X)$, baseline statistics $b$
\State \textbf{Output:} MLaaS-Aware Drift Exposure Score $S_{\text{MDES}}$
\vspace{3pt}
\State \textbf{Step 1: MLaaS Extraction Model Training}
\State Train lightweight extraction model $M'$ to approximate $M$: 
\Statex \hspace{1em}$M'(X) \approx M(X)$
\State Compute feature importance scores:
\Statex \hspace{1em}$I(x_i) = \frac{1}{N} \sum_{j=1}^{N} |M'(x_1^{(j)}, \dots, x_i^{(j)} + \delta, \dots, x_d^{(j)}) - M'(X^{(j)})|$
\State Form importance vector $\mathbf{I} = \{I(x_1), I(x_2), \dots, I(x_d)\}$
\State \textbf{Step 2: Compute Fréchet Data Drift Score}
%\State Estimate mean and covariance for baseline and current windows:
\Statex \hspace{1em}$\mu_b,\Sigma_b \gets \text{Stats}(X_b)$, \hspace{1em}$\mu_w,\Sigma_w \gets \text{Stats}(X_w)$
\Statex \hspace{1em}$S_{\text{FDDS}} = \lVert \mu_b - \mu_w \rVert_2^2 + 
\mathrm{Tr}\!\bigl(\Sigma_b + \Sigma_w - 2\sqrt{\Sigma_b\Sigma_w}\bigr)$
\State \textbf{Step 3: Compute MLaaS-Aware Drift Exposure Score}
\State Compute feature-weighted exposure:
\Statex \hspace{1em}$S_{\text{weighted}} = \sum_{i=1}^{d} I(x_i) \times S_{\text{FDDS}}$
\State Compute automated threshold:
\Statex \hspace{1em}$\theta_m = \mu + \beta \sigma$
\If{$S_{\text{weighted}} > \theta_m$}
    \State $S_{\text{MDES}} \gets 1$ \Comment{Real drift}
\Else
    \State $S_{\text{MDES}} \gets 0$ \Comment{Pseudo drift}
\EndIf
\end{algorithmic}
\end{algorithm}
\subsection{MLaaS Performance Drift Detection Model}
We propose a novel MPDD model to detect real and pseudo drift, specifically designed for IoT environments. It explicitly addresses two critical challenges: (i) detecting changes in dynamic input data properties, and (ii) monitoring concurrent changes in MLaaS behavior. \textit{Algorithm~\ref{alg:made}} describes the MLaaS Performance Drift Detection process. The inputs are the input data \(X_w\), the baseline window \(b\), and the current window \(w\). The output is the MLaaS-Aware Drift Exposure Score \(S_{\text{MDES}}\) (\textit{Algorithm~\ref{alg:made}}, Lines~1--2). The algorithm trains the MLaaS extraction model (\(M'\)) on input--output pairs \((X, Y)\) to approximate the underlying MLaaS model. This enables changes in input data to be mapped to shifts in the model’s decision patterns. It then computes feature-importance scores \(I(x_i)\) for each input characteristic and collects them as \(\mathbf{I} = \{I(x_1), \ldots, I(x_d)\}\) (\textit{Algorithm~\ref{alg:made}}, Lines~3--6). Next, the algorithm computes the Fr\'echet Data Drift Score (FDDS) to compare the current input statistics with the baseline statistics. This produces a single score that quantifies the distributional change (\textit{Algorithm~\ref{alg:made}}, Line~7). The MLaaS-Aware Drift Exposure Score (MDES) is then evaluated using Eq.~\ref{eq:MDES}, which combines \(\mathbf{I}\) with \(\mathrm{FDDS}\) to weight distributional shifts by feature influence (\textit{Algorithm~\ref{alg:made}}, Lines~8--10). Finally, the score is compared against a user-defined threshold \(\theta_m\) to classify the event as real drift (\(S_{\text{MDES}} = 1\)) or pseudo drift (\(S_{\text{MDES}} = 0\)) (\textit{Algorithm~\ref{alg:made}}, Lines~10--15), and the procedure returns \(S_{\text{MDES}}\).
\subsection{Adaptive-Temporal Performance Drift Detection Mechanism (APDDM)}
We propose an APDDM to adaptively adjust the window size $W_s$ based on recent drift patterns, ensuring stable performance-drift detection in MLaaS environments. The algorithm takes as inputs a stream of blocks $\{X_t\}$, buffer length $b$, initial window size $W_s$, drift ratio threshold $\theta_r$, trend length $L$, adaptation step $\Delta W$, and window bounds $[W_{\min}, W_{\max}]$ (\textit{Algorithm~\ref{alg:apddm}, line~1}), and outputs the detected drift triggers and final adapted window size (\textit{Algorithm~\ref{alg:apddm}, line~2}). The procedure initializes a buffer $\mathcal{B}$ to store recent drift scores along with two counters $c_u$ and $c_o$ that track consecutive drift and non-drift windows (\textit{Algorithm~\ref{alg:apddm}, lines~3--4}). For each incoming block $X_t$, the MPDD model computes a drift score $m_t$ and appends it to the buffer while retaining only the most recent observations (\textit{Algorithm~\ref{alg:apddm}, lines~4--6}). Once the buffer reaches capacity, the algorithm evaluates the proportion of scores exceeding the threshold to classify the current window as drift or stable (\textit{Algorithm~\ref{alg:apddm}, lines~7--8}). Accordingly, $c_u$ is incremented and $c_o$ reset under drift conditions, whereas stable states increment $c_o$ and reset $c_u$ (\textit{Algorithm~\ref{alg:apddm}, lines~9--12}), capturing short-term temporal trends in detection outcomes. If drift persists for at least $L$ consecutive windows, the method interprets this as under-sensitivity and reduces $W_s$ by $\Delta W$ within the allowable bounds (\textit{Algorithm~\ref{alg:apddm}, lines~14--15}). Conversely, prolonged stability for $L$ windows indicates over-sensitivity, leading to an increase in the window size to avoid unnecessary evaluations (\textit{Algorithm~\ref{alg:apddm}, lines~17--18}). After processing all blocks, the algorithm returns the final adapted window size (\textit{Algorithm~\ref{alg:apddm}, line~22}). Through this feedback-driven strategy, APDDM balances responsiveness and stability, reducing false alarms while ensuring timely detection of performance drift.
\begin{algorithm}[t]
\caption{Adaptive-Temporal Performance Drift Detection Mechanism (APDDM)}
\label{alg:apddm}
\begin{algorithmic}[1]
\State \textbf{Input:} \( \langle \{X_t\}, b, W_s, \theta_r, L, \Delta W, W_{min}, W_{max} \rangle \)
\State \textbf{Output:} drift triggers, \(W_s\)
\State \(\mathcal{B}\gets[\ ]\), \(c_u\gets0\), \(c_o\gets0\)
\For{each block \(X_t\)}
    \State \(m_t \gets \textsc{MPDD}(X_t,b)\) \Comment{\(m_t \in \{0,1\}\), where 1 indicates real drift}
    \State append \(m_t\) to \(\mathcal{B}\); if \(|\mathcal{B}|>W_s\), remove oldest
    \If{\(|\mathcal{B}|=W_s\)}
        \State \(r \gets \frac{1}{W_s}\sum_{m\in\mathcal{B}} m\) \Comment{drift ratio in recent window}
        \If{\(r>\theta_r\)}
            \State trigger drift; \(c_u\gets c_u+1\); \(c_o\gets0\)
        \Else
            \State \(c_o\gets c_o+1\); \(c_u\gets0\)
        \EndIf
        \If{\(c_u \ge L\)}
            \State \(W_s\gets \max(W_{min},\,W_s-\Delta W)\); \(c_u\gets0\)
        \EndIf
        \If{\(c_o \ge L\)}
            \State \(W_s\gets \min(W_{max},\,W_s+\Delta W)\); \(c_o\gets0\)
        \EndIf
    \EndIf
\EndFor
\State \Return drift triggers, \(W_s\)
\end{algorithmic}
\end{algorithm}

\section{Computational complexity analysis}
The computational complexity of Algorithm 3 (APDDM) is primarily governed by the per-block drift score computation and the adaptive window management operations. For each incoming data block $X_t$ of size $n$ with $d$ features, the algorithm invokes the MPDD scoring function once and then performs lightweight buffer updates and ratio evaluation. Assuming the extraction model and baseline statistics are precomputed offline (i.e., no online retraining), the online MPDD computation involves feature-wise statistical comparison and weighted aggregation, requiring a single pass over the block and yielding a time complexity of $O(nd)$ under a diagonal/statistical representation, or up to $O(nd^2)$ if full covariance-based distance measures are employed. The APDDM control logic consists of inserting the drift score into a bounded buffer of size $b$, computing the drift ratio over the buffer, and updating counters and window size, resulting in $O(W_s)$ per block (or $O(1)$ if a running count is maintained). Therefore, the overall per-block complexity of the algorithm can be expressed as $O(nd + W_s)$, and over $T$ processed blocks, the total time complexity of the proposed adaptive drift prediction mechanism is given by
\begin{equation}
\mathcal{T}(T) = O\!\left(\sum_{t=1}^{T} \left(nd + W_s(t)\right)\right)
\end{equation}
which simplifies to $O(Tnd)$ in typical streaming settings where $W_s \ll n$. The space complexity is modest, requiring $O(W_s)$ memory to store the adaptive drift-score buffer and $O(d)$ (or $O(d^2)$ for full covariance) for maintaining baseline statistical summaries, making the algorithm computationally efficient and scalable for continuous MLaaS monitoring scenarios without incurring any retraining overhead.
\vspace{-3mm}
\section{Experiments and Results}
In this section, we evaluate the proposed MPDD for IoT environments. First, we assess the effectiveness of the MPDD model in distinguishing real drift from pseudo drift by comparing it with representative baseline techniques, including ADWIN, MMD, D3, STUDD, and SCSD, using accuracy, precision, and recall. Second, we evaluate the timeliness of the proposed APDDM across multiple datasets. This experiment investigates the effectiveness of adaptive temporal monitoring in improving drift responsiveness compared with fixed-interval monitoring and baseline drift detectors. Performance is evaluated using detection accuracy, accuracy gain, miss detection ratio, and false negatives. Third, we evaluate the MLaaS extraction model by analyzing its fidelity and predictive behavior relative to the original black-box MLaaS models. In addition, we assess the impact of surrogate approximation on the overall detection performance. The MLaaS servies are extracted from the MDA data generation framework \cite{kanneganti2026machine}. All experiments were conducted on an Intel Core i7 machine with 16 GB RAM using Python. The source code is publicly available in the repository\footnote{\url{https://anonymous.4open.science/r/MPDD_IoT-6D16/README}}.
\vspace{-3mm}

\subsection{Experiment Setup and Dataset}
In this section, we discuss the datasets, drift setup, and baseline techniques used to evaluate the proposed framework.
\subsubsection{Dataset}
\begin{itemize}[itemsep=0ex, leftmargin=2ex, label=\textbullet]
\item \textit{Human Activity Recognition (HAR) \cite{misc_pamap2_physical_activity_monitoring_231}:} We use the PAMAP2 dataset containing 3,850,505 samples across 52 sensor channels collected from multiple daily activities. 
\item \textit{Electricity \cite{electricityloaddiagrams20112014_321}:} The Electricity dataset has 45,325 samples with nine features from the New South Wales electricity market, capturing price changes every five minutes.
\item \textit{Weather \cite{noaa_gsod_kaggle_2025}:} The NOAA Weather dataset includes 96,454 records with nine features describing weather attributes.
\item \textit{Airline \cite{DVN/HG7NV7_2008}:} The Airline Delay dataset contains 539,395 instances with 8 features, including arrival and departure records of U.S. commercial flights.
\item \textit{Poker \cite{cattral2002poker}:} The Poker dataset consists of 1,000,000 samples with 11 attributes, where each instance represents a hand of five cards encoded by rank and suit.
\end{itemize}

\subsubsection{Drift Data Generation}

Traditional studies simulate different forms of concept drift, including sudden, gradual, incremental, and recurrent changes in data streams. To comprehensively evaluate the proposed framework, we generate representative drift scenarios covering all four drift categories using the widely adopted \textit{SEA generator} \cite{pedregosa2011scikit}. The generated scenarios introduce controlled changes in the data distribution, allowing the effectiveness of the framework to be assessed under diverse drift behaviors without being restricted to a specific drift type. Examples of the generated drift scenarios across datasets are provided in \textit{Appendix B}.

\subsubsection{Baseline Techniques}
Existing literature does not explicitly address the challenge of MLaaS performance drift, particularly in black-box IoT settings. However, several related drift detection methods partially capture distributional or model-related changes and can serve as reasonable baselines for comparison. To ensure a fair and consistent evaluation, we compare the proposed MPDD framework with established benchmark techniques, including ADWIN~\cite{bifet2007learning} and MMD~\cite{sinn2012detecting}, which represent error-based and statistical data-driven drift detection approaches. In addition, we include representative black-box unsupervised methods that align with realistic MLaaS deployment constraints, namely STUDD~\cite{cerqueira2023studd}, SCSD~\cite{jang2022sequential}, and D3~\cite{gozuaccik2019unsupervised}. All baseline models are implemented using their recommended drift thresholds and evaluated consistently across the selected datasets.

\subsubsection{Configuration of the MLaaS Extraction Model}
The MLaaS extraction model ($M'$) plays a key role in the proposed framework by approximating the behaviour of the black-box MLaaS service. This model enables the extraction of feature-importance information used for performance drift detection. Table~\ref{tab:extraction_configuration} summarizes the architecture and parameter configuration of the extraction model, including the decision-tree structure, splitting criterion, training target, and model settings. Table~\ref{tab:extraction_training_details} presents the training-data configuration used to construct $M'$, including the query-pool size, clean and drift sample composition, and train--test split across datasets. These details provide additional transparency regarding the design and training process of the extraction model.

\begin{table}[H]
\centering
\caption{Configuration of the MLaaS extraction model $M'$.}
\label{tab:extraction_configuration}
\scriptsize
\renewcommand{\arraystretch}{1.05}
\setlength{\tabcolsep}{4pt}
\resizebox{\columnwidth}{!}{%
\begin{tabular}{llll}
\hline
\textbf{Parameter} & \textbf{Configuration} &
\textbf{Parameter} & \textbf{Configuration} \\
\hline
Model & CART decision tree &
Criterion & Gini impurity \\

Split strategy & Best split &
Maximum depth & 4 \\

Min samples split & 2 &
Min samples leaf & 1 \\

Maximum features & All variables &
Training target & MLaaS labels \\

Train/Test split & 50\% / 50\% &
Training data & 60\% clean + 40\% drift \\
\hline
\end{tabular}%
}
\end{table}

\begin{table}[H]
\centering
\caption{Training data configuration of the MLaaS extraction model}
\label{tab:extraction_training_details}
\scriptsize
\renewcommand{\arraystretch}{1.05}
\setlength{\tabcolsep}{4pt}
\begin{tabular}{lrrrrr}
\hline
\textbf{Dataset} &
\textbf{Query Pool} &
\textbf{Clean} &
\textbf{Drift} &
\textbf{Train} &
\textbf{Test} \\
\hline
HAR         & 100,000 & 60,000 & 40,000 & 50,000 & 50,000 \\
Electricity & 75,000  & 45,000 & 30,000 & 37,500 & 37,500 \\
Weather     & 160,000 & 96,000 & 64,000 & 80,000 & 80,000 \\
Poker       & 160,000 & 96,000 & 64,000 & 80,000 & 80,000 \\
Airline     & 16,649  & 9,989  & 6,660  & 8,324  & 8,325 \\
\hline
\end{tabular}
\end{table}

\begin{table*}[t]
\centering
\caption{Comparison of MLaaS Performance Drift Detection (MPDD) under pseudo drift and real drift across datasets.}
\label{tab:framework_comparison}

\scriptsize
\renewcommand{\arraystretch}{1.15}
\setlength{\tabcolsep}{3.2pt}

\resizebox{\textwidth}{!}{%
\begin{tabular}{ll ccc ccc ccc ccc ccc}
\toprule
\multirow{2}{*}{\textbf{Method}} &
\multirow{2}{*}{\textbf{Drift}} &
\multicolumn{3}{c}{\textbf{HAR}\cite{misc_pamap2_physical_activity_monitoring_231}} &
\multicolumn{3}{c}{\textbf{Electricity}\cite{electricityloaddiagrams20112014_321}} &
\multicolumn{3}{c}{\textbf{Weather}\cite{noaa_gsod_kaggle_2025}} &
\multicolumn{3}{c}{\textbf{Poker}\cite{cattral2002poker}} &
\multicolumn{3}{c}{\textbf{Airline}\cite{DVN/HG7NV7_2008}} \\
\cmidrule(lr){3-5}\cmidrule(lr){6-8}\cmidrule(lr){9-11}
\cmidrule(lr){12-14}\cmidrule(lr){15-17}
& & \textbf{Acc} & \textbf{Prec} & \textbf{Rec}
  & \textbf{Acc} & \textbf{Prec} & \textbf{Rec}
  & \textbf{Acc} & \textbf{Prec} & \textbf{Rec}
  & \textbf{Acc} & \textbf{Prec} & \textbf{Rec}
  & \textbf{Acc} & \textbf{Prec} & \textbf{Rec} \\
\midrule

\multirow{2}{*}{\textbf{MPDD}}
& \textbf{Pseudo}
& \multirow{2}{*}{\textbf{0.91}} & 0.88 & 0.98
& \multirow{2}{*}{\textbf{0.90}} & 0.87 & 0.98
& \multirow{2}{*}{\textbf{0.89}} & 0.85 & 0.98
& \multirow{2}{*}{\textbf{0.88}} & 0.89 & 0.94
& \multirow{2}{*}{\textbf{0.90}} & 0.87 & 0.98\\
& \textbf{Real}
& & 0.98 & 0.71
& & 0.99 & 0.72
& & 0.98 & 0.71
& & 0.85 & 0.75
& & 0.98 & 0.69 \\
\midrule

\multirow{2}{*}{ADWIN~\cite{bifet2007learning}}
& \textbf{Pseudo}
& \multirow{2}{*}{0.62} & 0.66 & 0.74
& \multirow{2}{*}{0.77} & 0.77 & 0.83
& \multirow{2}{*}{0.50} & 0.46 & 0.65
& \multirow{2}{*}{0.54} & 0.61 & 0.64
& \multirow{2}{*}{0.65} & 0.71 & 0.71 \\
& \textbf{Real}
& & 0.53 & 0.44
& & 0.77 & 0.69
& & 0.57 & 0.38
& & 0.42 & 0.39
& & 0.56 & 0.56 \\
\midrule

\multirow{2}{*}{MMD~\cite{sinn2012detecting}}
& \textbf{Pseudo}
& \multirow{2}{*}{0.52} & 0.59 & 0.64
& \multirow{2}{*}{0.69} & 0.69 & 0.82
& \multirow{2}{*}{0.54} & 0.49 & 0.70
& \multirow{2}{*}{0.54} & 0.61 & 0.65
& \multirow{2}{*}{0.65} & 0.71 & 0.71 \\
& \textbf{Real}
& & 0.39 & 0.34
& & 0.71 & 0.53
& & 0.62 & 0.40
& & 0.42 & 0.38
& & 0.56 & 0.56 \\
\midrule

\multirow{2}{*}{D3~\cite{gozuaccik2019unsupervised}}
& \textbf{Pseudo}
& \multirow{2}{*}{0.46} &0.98 & 0.10
& \multirow{2}{*}{0.72} & 0.81 & 0.64
& \multirow{2}{*}{0.57} & 0.51 & 0.74
& \multirow{2}{*}{0.53} & 0.59 & 0.69
& \multirow{2}{*}{0.60} & 0.60 & 0.98 \\
& \textbf{Real}
& & 0.43 & 0.98
& & 0.65 & 0.82
& & 0.67 & 0.43
& & 0.38 & 0.28
& & 0.10 & 0.10 \\
\midrule

\multirow{2}{*}{STUDD~\cite{cerqueira2023studd}}
& \textbf{Pseudo}
& \multirow{2}{*}{0.69} & 0.69 & 0.99
& \multirow{2}{*}{0.66} & 0.66 & 0.98
& \multirow{2}{*}{0.61} & 0.62 & 0.95
& \multirow{2}{*}{0.68} & 0.68 & 0.98
& \multirow{2}{*}{0.68} & 0.68 & 0.98 \\
& \textbf{Real}
& & 0.68 & 0.04
& & 0.57 & 0.05
& & 0.37 & 0.05
& & 0.10 & 0.10
& & 0.10 & 0.10 \\
\midrule

\multirow{2}{*}{SCSD~\cite{jang2022sequential}}
& \textbf{Pseudo}
& \multirow{2}{*}{0.65} & 0.99 & 0.50
& \multirow{2}{*}{0.56} & 0.98 & 0.33
& \multirow{2}{*}{0.69} & 0.67 & 0.99
& \multirow{2}{*}{0.70} & 0.70 & 0.98
& \multirow{2}{*}{0.41} & 0.98 & 0.15 \\
& \textbf{Real}
& & 0.48 & 0.98
& & 0.44 & 0.98
& & 0.95 & 0.21
& & 0.97 & 0.06
& &0.35 & 0.90 \\
\bottomrule
\end{tabular}%
}
\end{table*}

\begin{figure*}[t]
\centering

% -------- Row 1 --------
\begin{minipage}{0.32\textwidth}
\centering
\includegraphics[width=\linewidth]{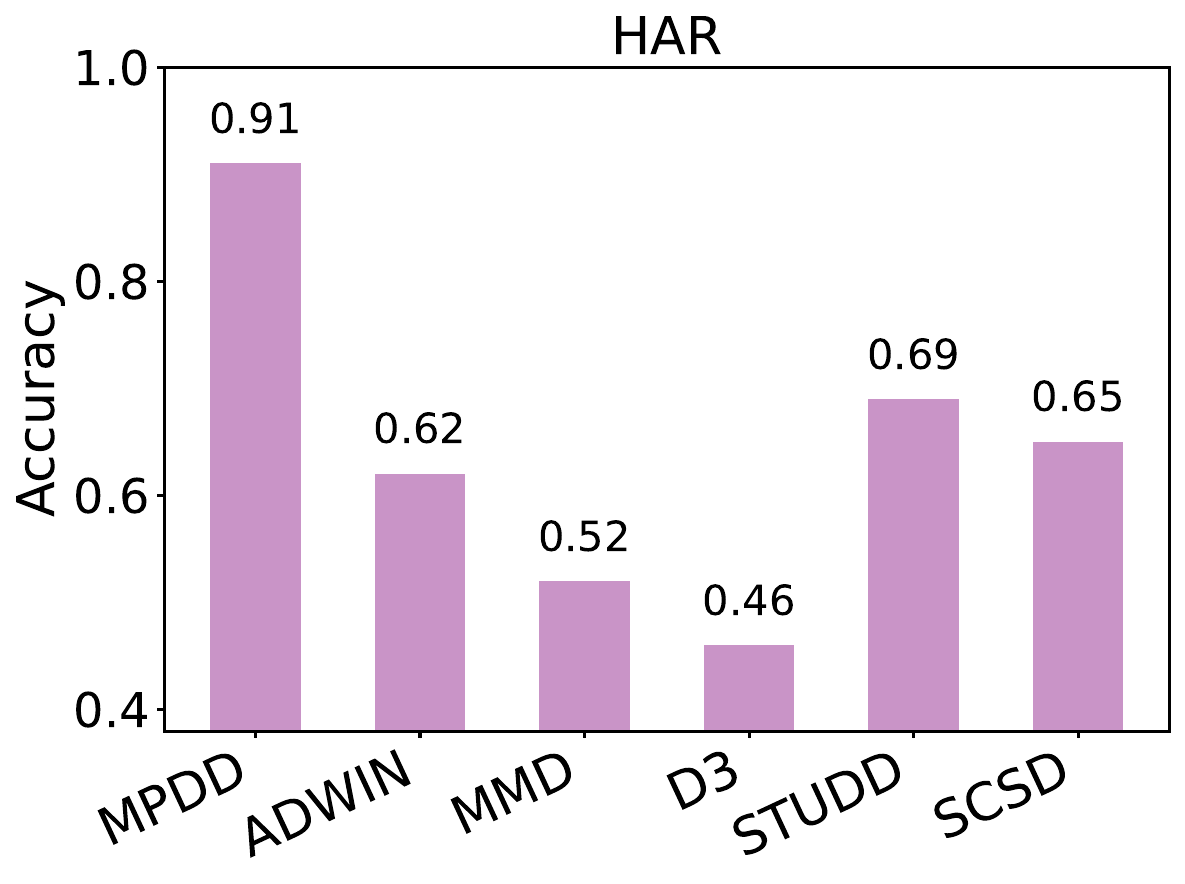}
\\ {\small (a)}
\end{minipage}\hfill
\begin{minipage}{0.32\textwidth}
\centering
\includegraphics[width=\linewidth]{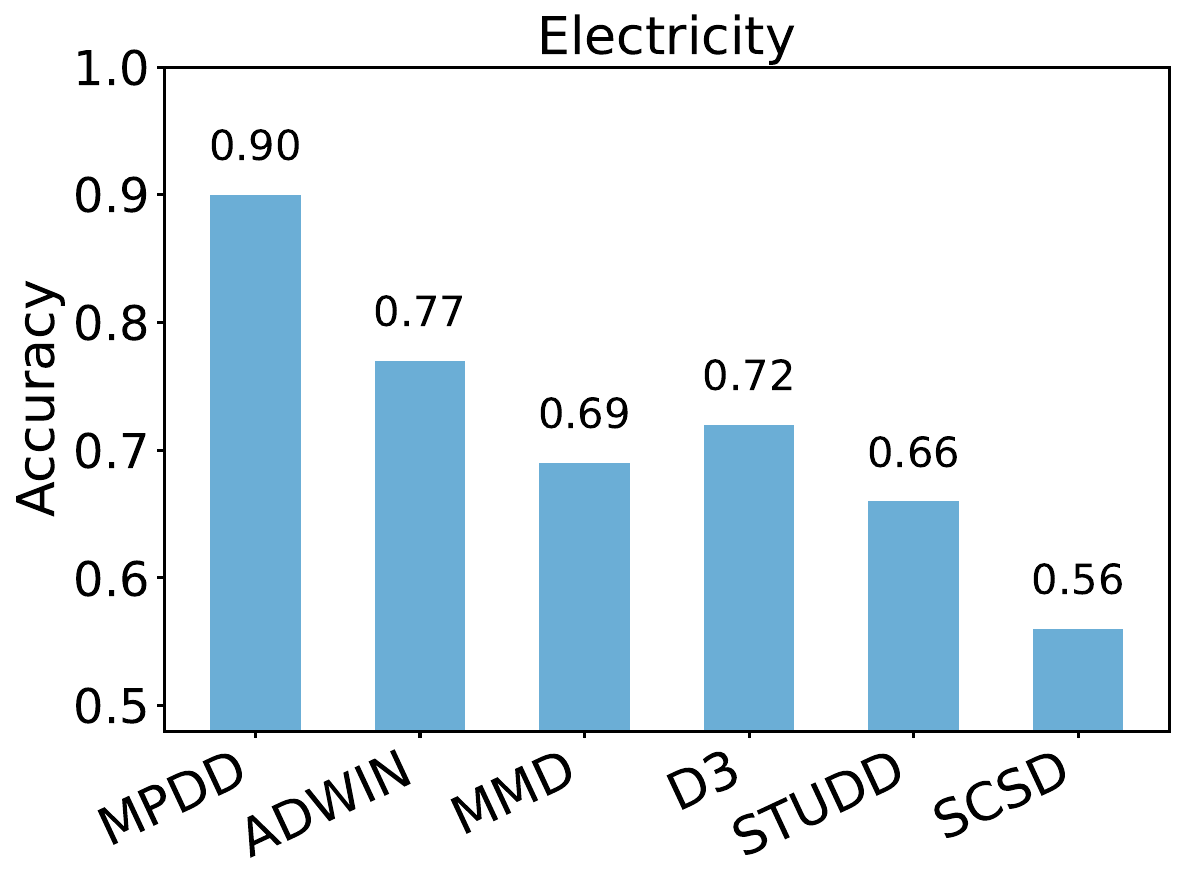}
\\ {\small (b)}
\end{minipage}\hfill
\begin{minipage}{0.32\textwidth}
\centering
\includegraphics[width=\linewidth]{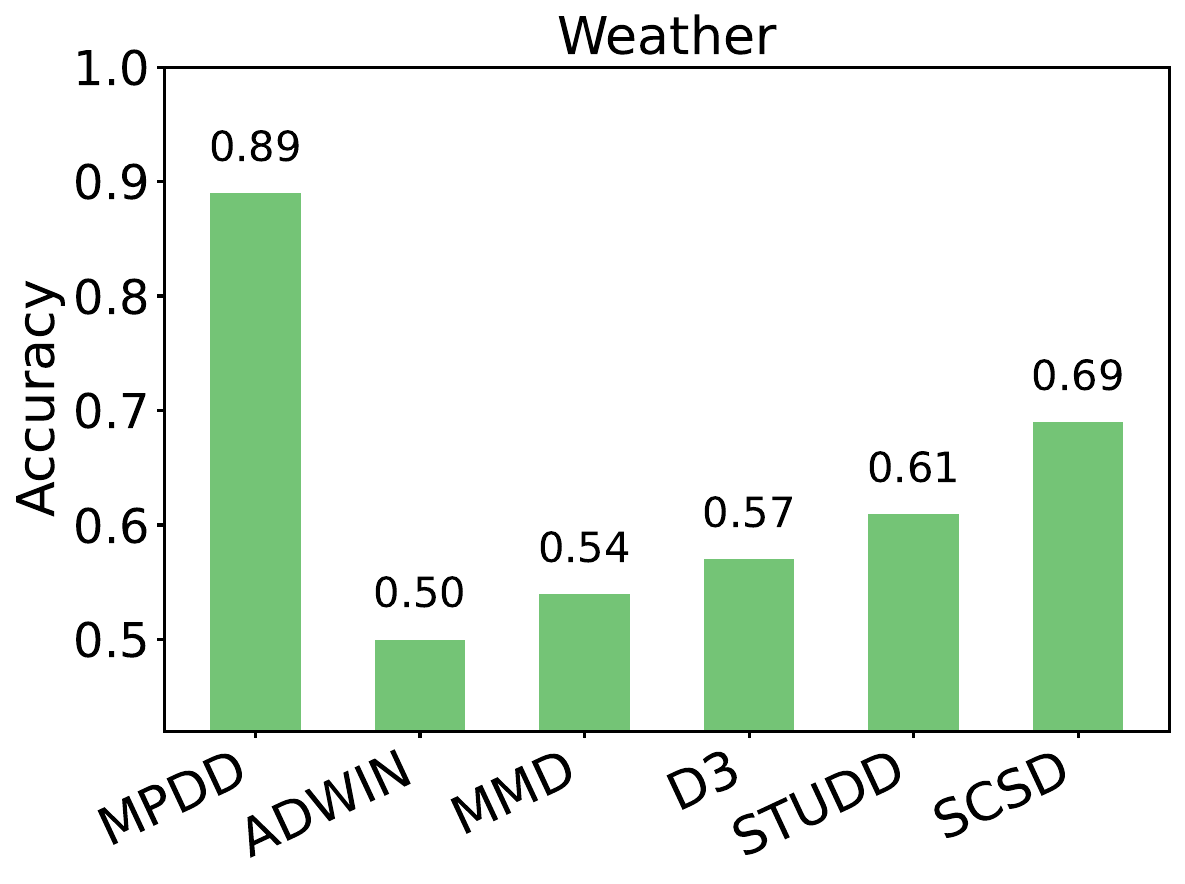}
\\ {\small (c)}
\end{minipage}

\vspace{3mm}

% -------- Row 2 --------
\begin{minipage}{0.32\textwidth}
\centering
\includegraphics[width=\linewidth]{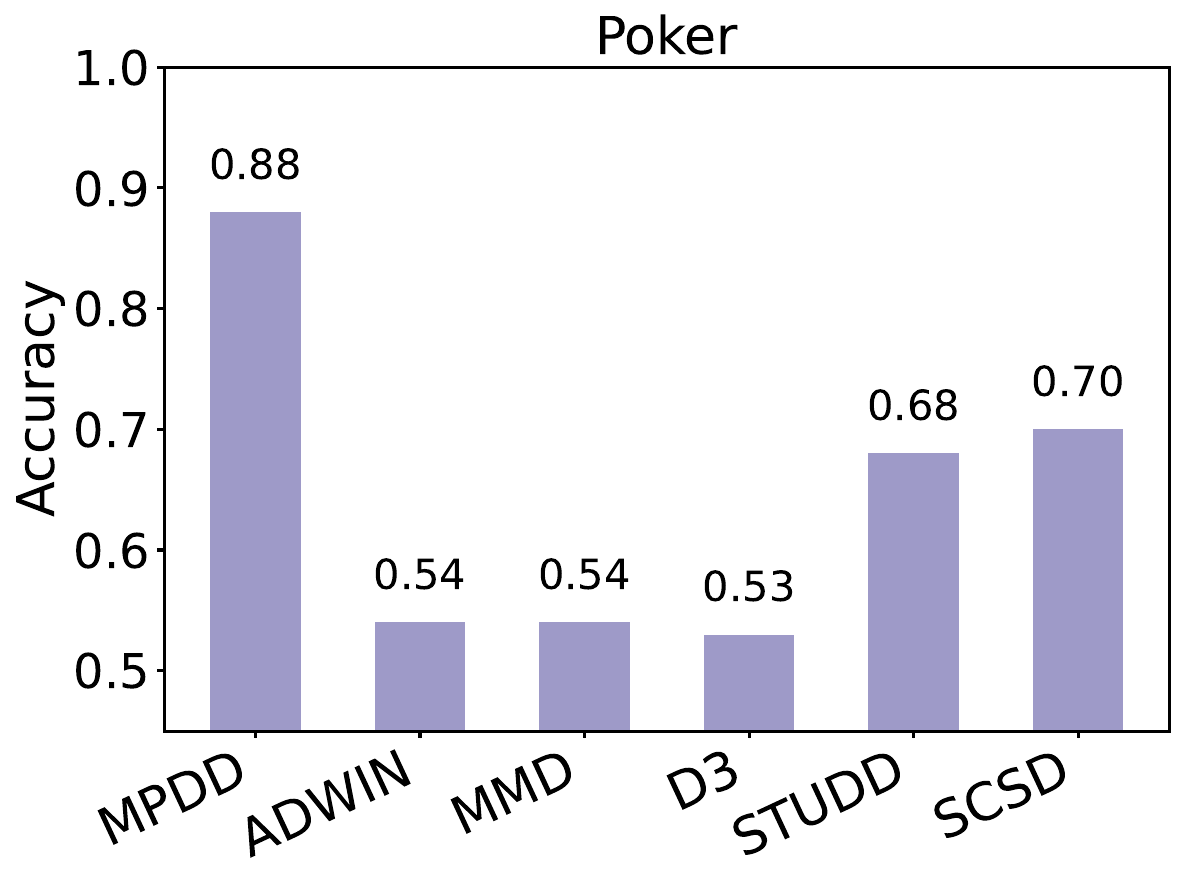}
\\ {\small (d)}
\end{minipage}\hfill
\begin{minipage}{0.32\textwidth}
\centering
\includegraphics[width=\linewidth]{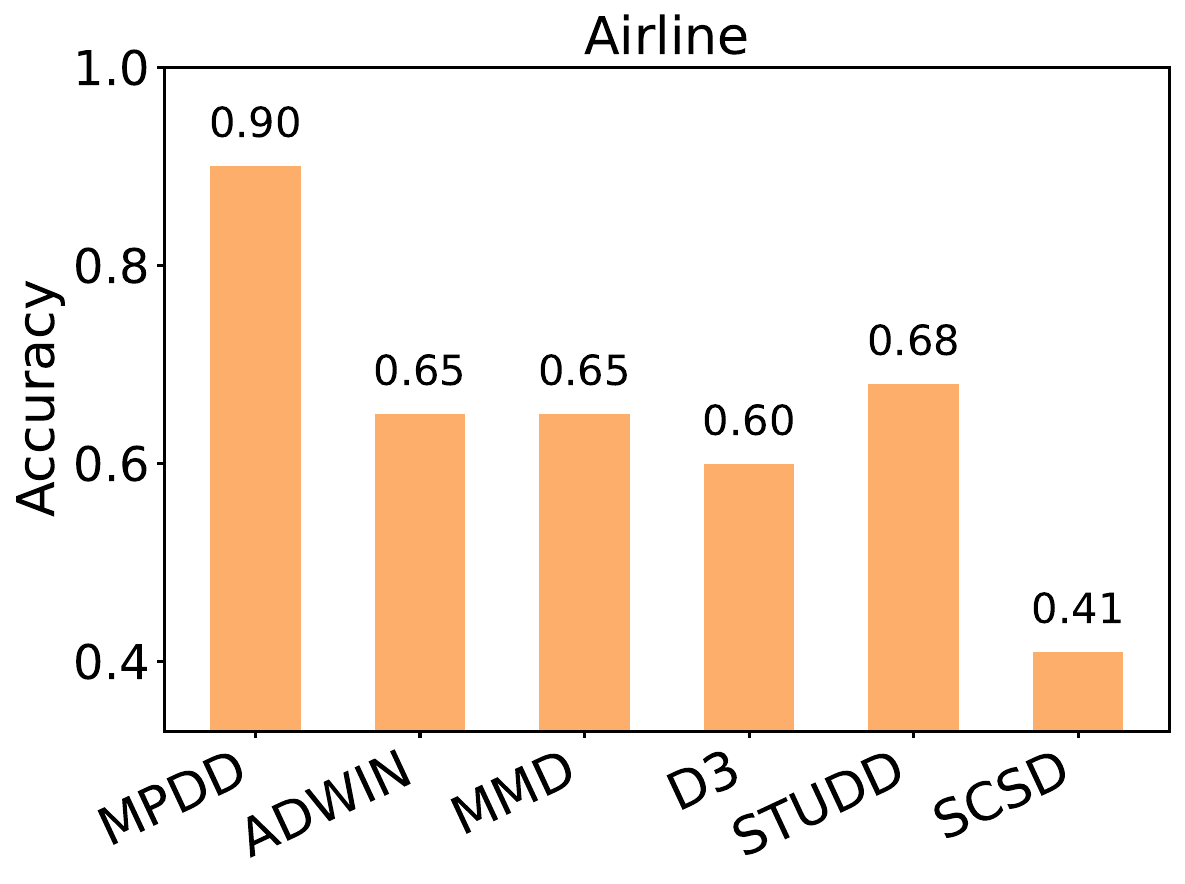}
\\ {\small (e)}
\end{minipage}\hfill
\begin{minipage}{0.32\textwidth}
\centering
\includegraphics[width=\linewidth]{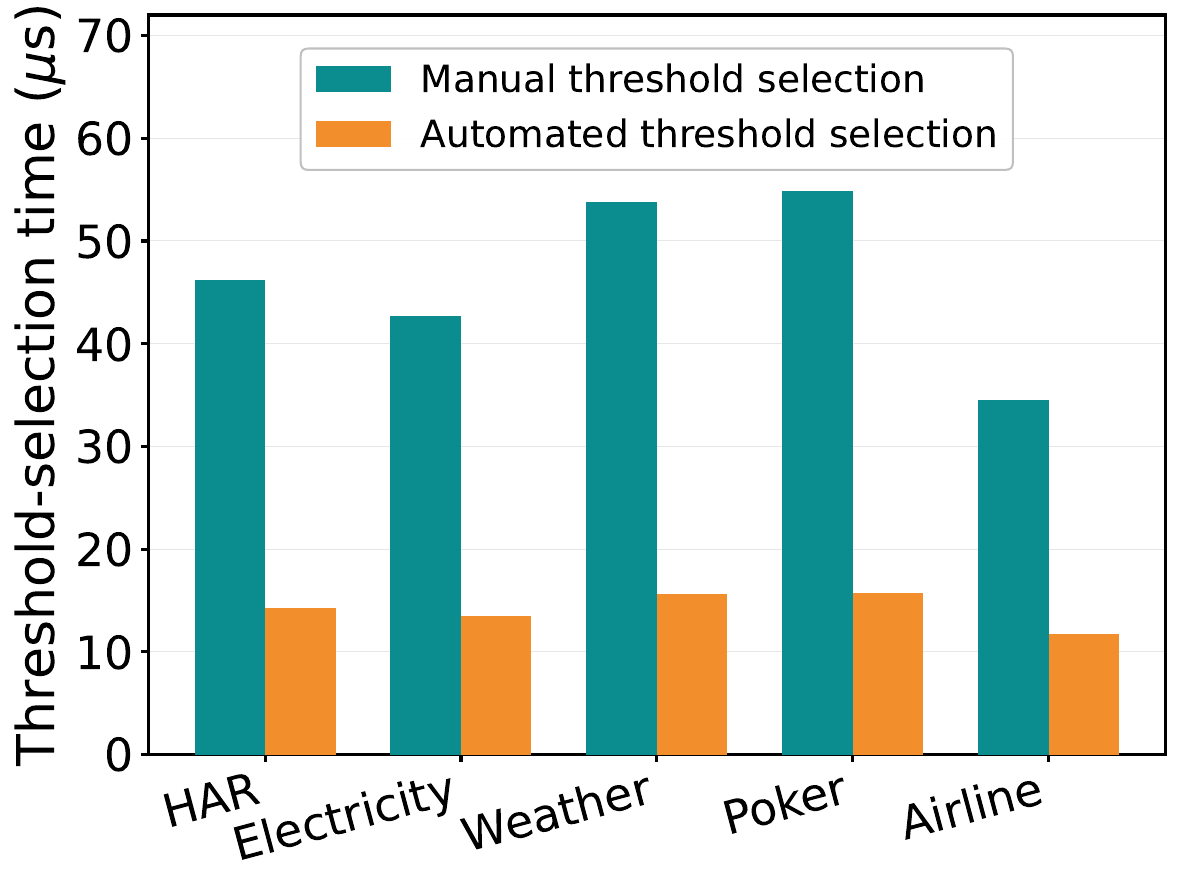}
\\ {\small (f)}
\end{minipage}
\vspace{3mm}
\normalsize
\caption{\normalsize Comparative accuracy analysis of drift detection techniques across datasets: (a) HAR, (b) Electricity, (c) Weather, (d) Poker, (e) Airline, and (f) threshold sensitivity analysis, evaluated using the proposed  MPDD model.}
\label{fig:accuracy_3col_2row}
\vspace{-2mm}
\end{figure*}

\subsection{Experiment 1: Evaluation of MLaaS Performance Drift Detection approach}
The proposed MPDD model aims to detect real and pseudo drift in IoT environments. To evaluate its effectiveness, we use standard metrics including accuracy, precision, and recall. \textit{Table~\ref{tab:framework_comparison}} compares MPDD against benchmark drift detection techniques, including ADWIN, MMD, STUDD, SCSD, and D3, across multiple datasets. Identifying an appropriate threshold to distinguish real and pseudo drift is a critical challenge in drift detection. Since trial-and-error threshold selection may not generalize well across different IoT environments, we adopt an automated threshold adjustment mechanism inspired by DDM and EDDM, where the drift-exposure threshold ($\theta_m$) is automatically determined from the observed drift-exposure scores. Table~\ref{tab:framework_comparison} shows that MPDD consistently achieves the highest overall detection accuracy across all datasets while maintaining balanced precision and recall for both real and pseudo drift scenarios. For pseudo drift, MPDD achieves accuracy values of 0.91 on HAR, 0.90 on Electricity, 0.89 on Weather, 0.88 on Poker, and 0.90 on Airline, with precision ranging from 0.85 to 0.89 and recall values consistently above 0.94. More importantly, MPDD maintains strong performance under real drift conditions, where competing approaches often degrade substantially. Across all datasets, it achieves precision values of up to 0.99 and recall values between 0.69 and 0.75, demonstrating reliable identification of genuine MLaaS performance degradation. In contrast, traditional data-distribution-based methods such as ADWIN and MMD show lower and less balanced performance. While they achieve moderate pseudo-drift detection in some datasets, their real-drift recall ranges from only 0.34 to 0.69. This suggests that methods relying primarily on data distribution changes may fail when variations in the input data do not directly affect MLaaS prediction behaviour, indicating that data drift does not always translate to performance drift. Black-box unsupervised approaches such as D3, STUDD, and SCSD exhibit mixed performance. D3 achieves competitive results on some datasets, such as Electricity with a real-drift recall of 0.82, but shows severe precision-recall imbalance on others, including HAR where pseudo-drift recall falls to 0.10. Similarly, STUDD achieves high pseudo-drift recall across all datasets (0.95--0.99) but performs poorly for real drift detection, with recall values dropping to as low as 0.04--0.10. SCSD achieves high precision or recall for specific drift classes, but suffers from substantial precision-recall imbalance, resulting in inconsistent detection performance. Overall, these results confirm the robustness and generalization of MPDD in detecting both pseudo and real drift while maintaining a more balanced performance than existing data-based and black-box unsupervised drift detection approaches.

\textit{Figure~\ref{fig:accuracy_3col_2row}} compares the detection accuracy of MPDD and baseline methods across all datasets. Figures~\ref{fig:accuracy_3col_2row}(a)--(e) show that MPDD consistently achieves the highest accuracy on HAR, Electricity, Weather, Poker, and Airline. The performance gap between MPDD and the strongest competing method remains substantial across all datasets, highlighting the effectiveness of incorporating both feature-importance variations and data-drift characteristics into the drift-detection process. These results further demonstrate that MPDD maintains robust and consistent performance across diverse IoT environments and drift scenarios. Figure~\ref{fig:accuracy_3col_2row}(f) evaluates the efficiency of the automated threshold-selection mechanism. The results show that the proposed approach determines suitable drift-exposure thresholds significantly faster than the trial-and-error method, achieving an approximately threefold reduction in threshold-selection time. In addition to reducing computational overhead, the automated mechanism eliminates the need for dataset-specific threshold tuning for new IoT environments.

\subsection{Experiment 2: Timeliness Evaluation of Adaptive-Temporal Performance Drift Detection Mechanism (APDDM)}

\begin{figure}[!t]
\centering

\includegraphics[width=\columnwidth]{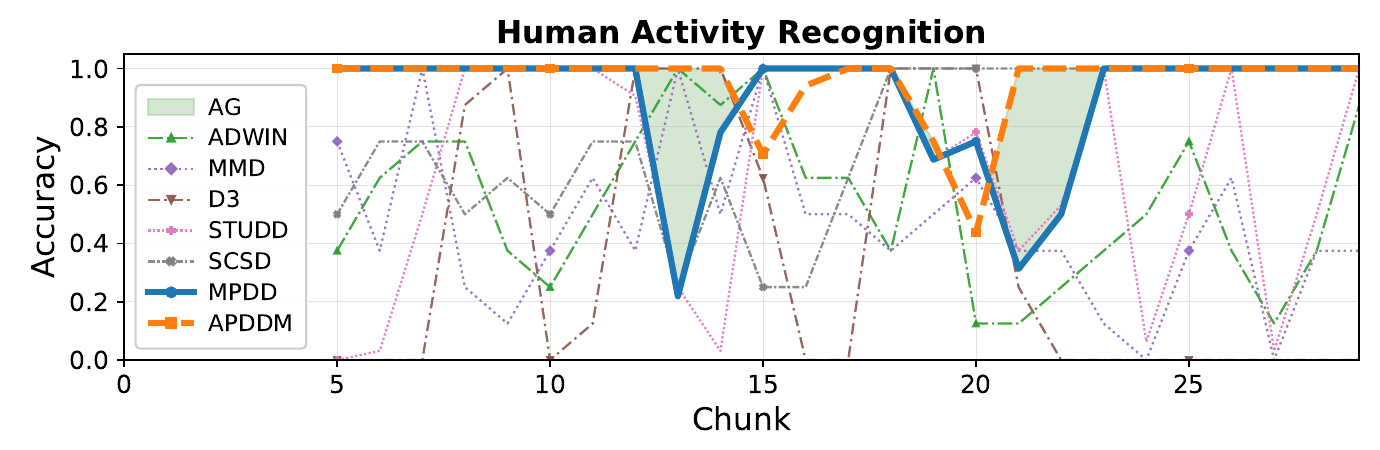}\\[-1mm]
{\small (a)}\\[0.5mm]

\includegraphics[width=\columnwidth]{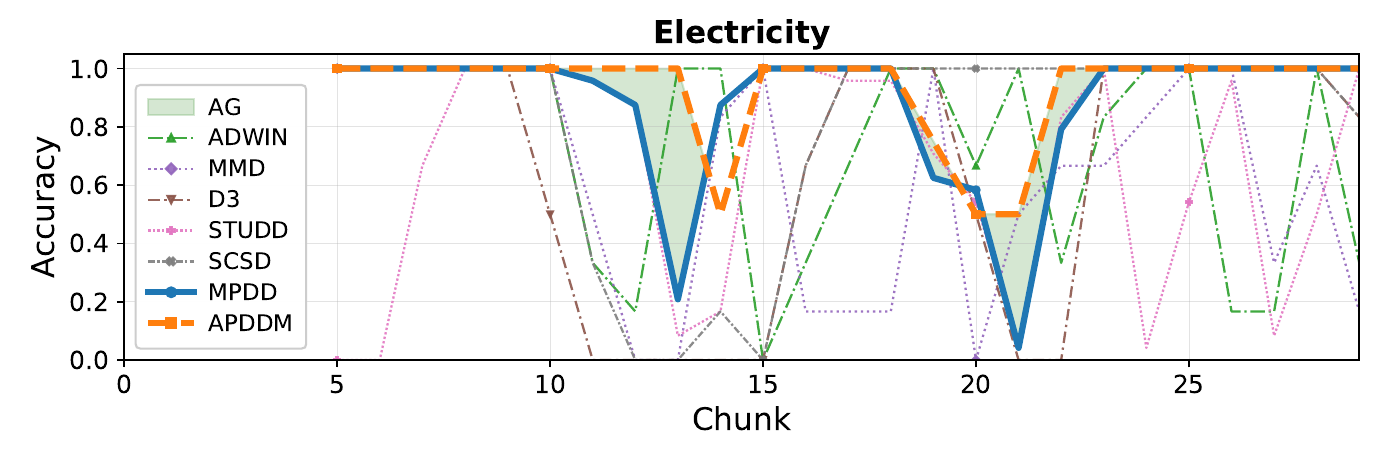}\\[-1mm]
{\small (b)}\\[0.5mm]

\includegraphics[width=\columnwidth]{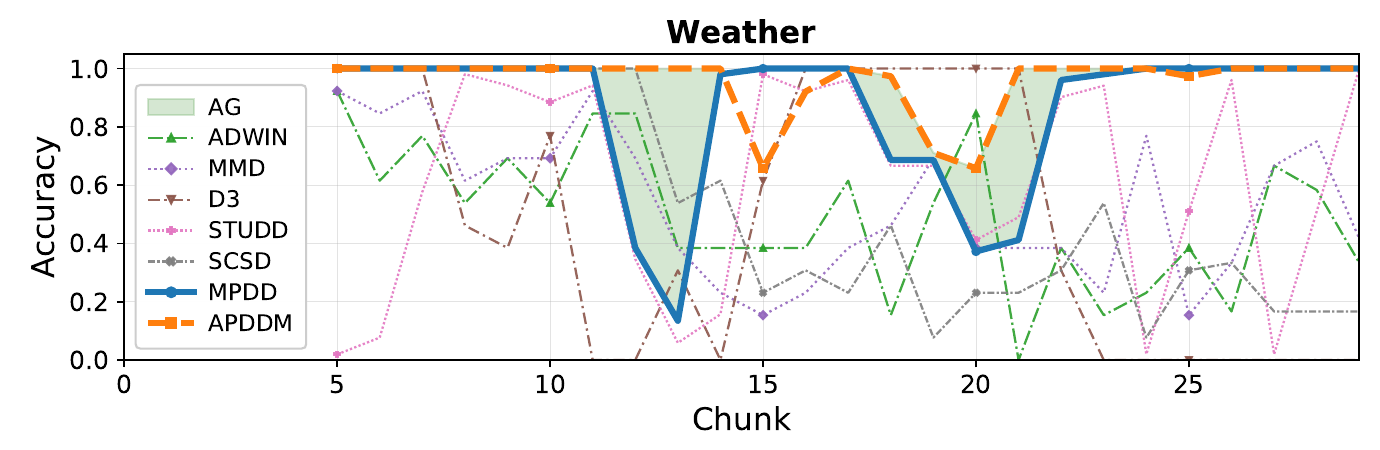}\\[-1mm]
{\small (c)}\\[0.5mm]

\includegraphics[width=\columnwidth]{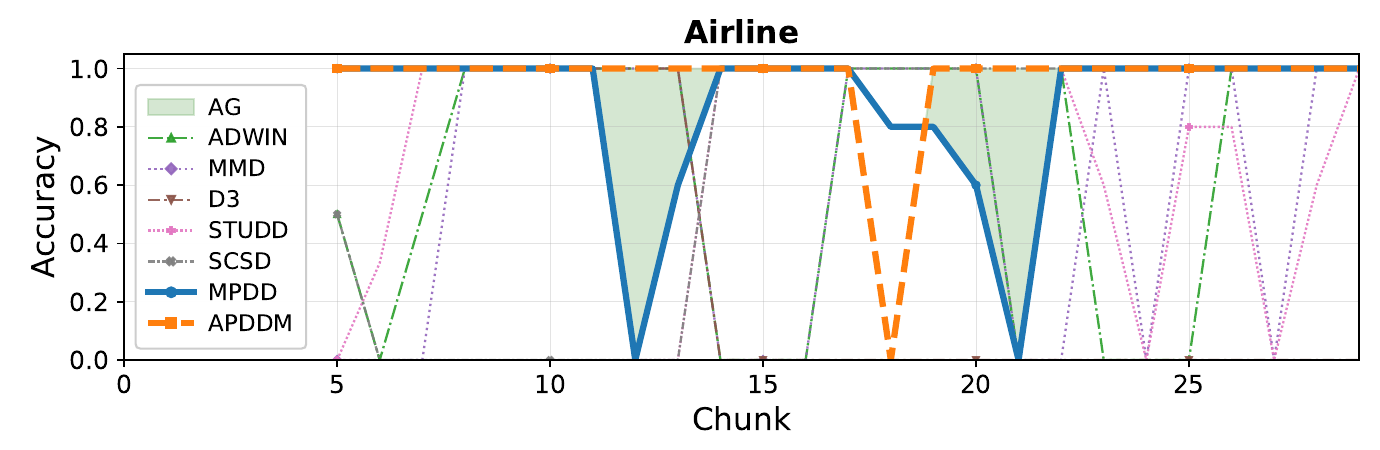}\\[-1mm]
{\small (d)}\\[0.5mm]

\includegraphics[width=\columnwidth]{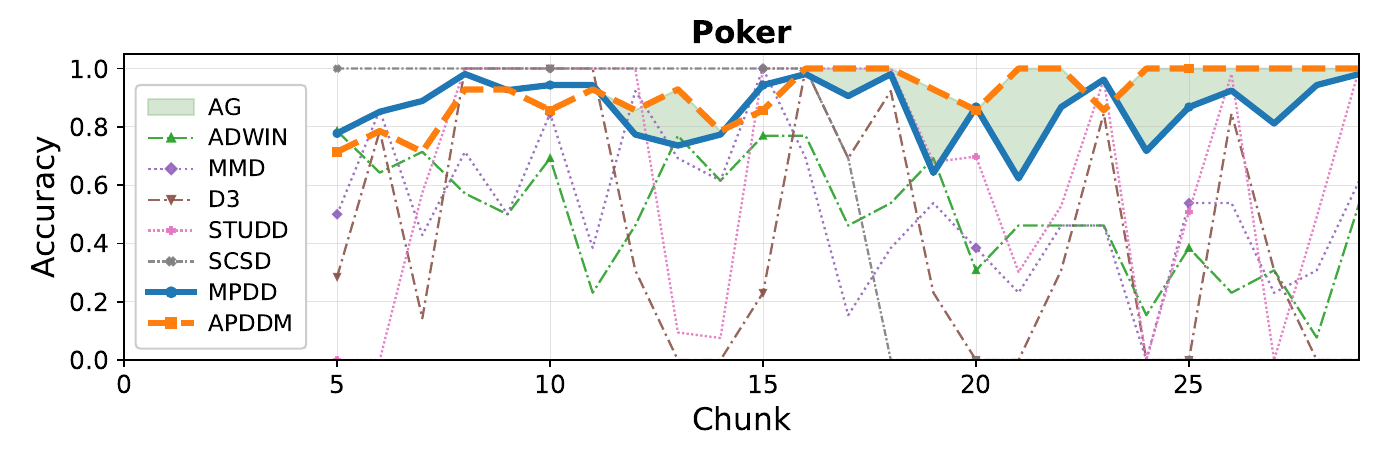}\\[-1mm]
{\small (e)}

\caption{\normalsize Timeliness evaluation of APDDM across datasets: (a) HAR, (b) Electricity, (c) Weather, (d) Airline, and (e) Poker, showing accuracy comparisons for each scenario.}
\label{fig:fn}
\vspace{-3mm}

\end{figure}

\begin{figure}[!t]
\centering
\includegraphics[width=\columnwidth]{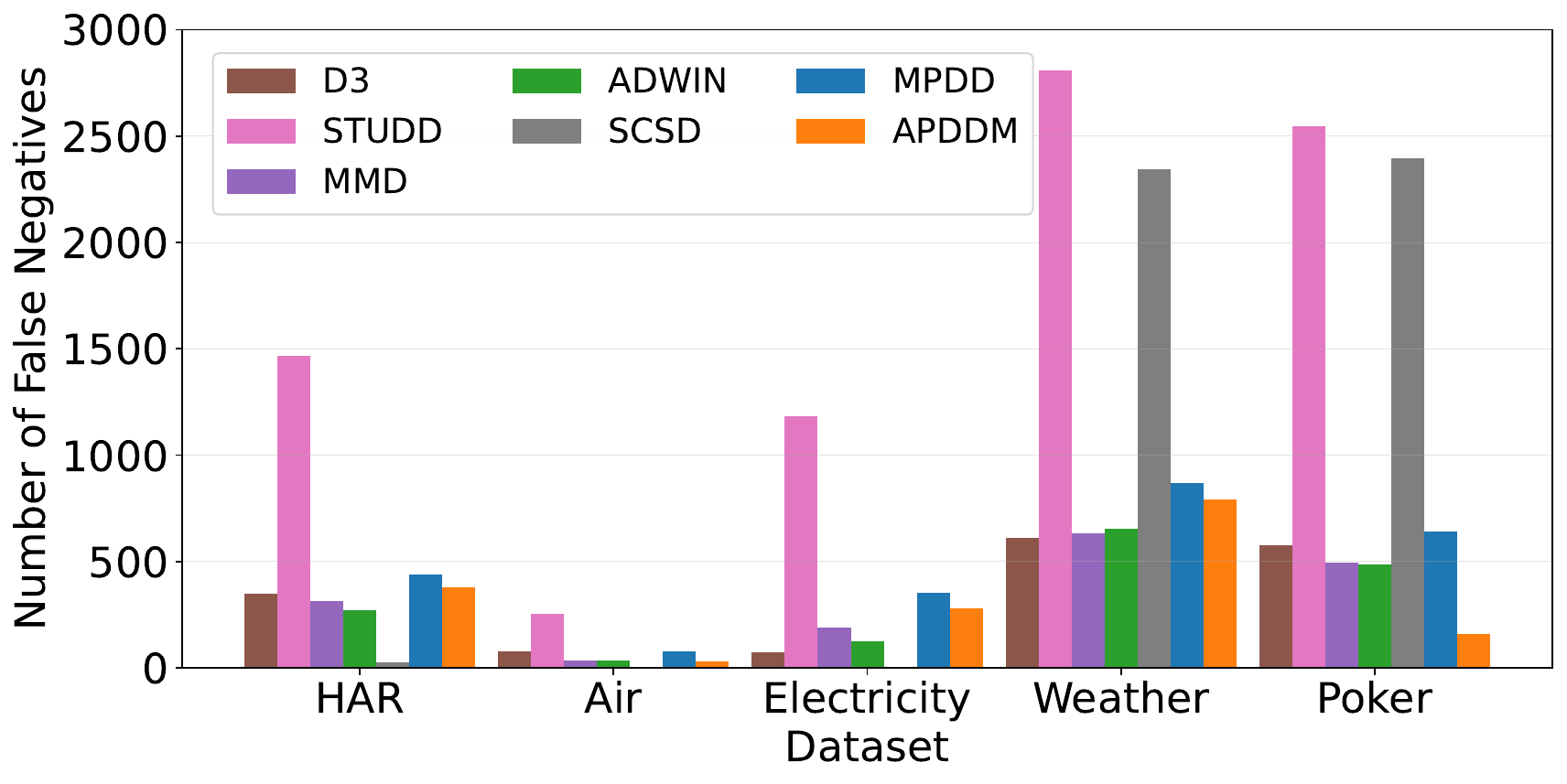}
\caption{\normalsize False negative count comparison across the HAR, Airline, Electricity, Weather, and Poker datasets.}
\label{fig:false_negative}
\vspace{-3mm}
\end{figure}

\textit{Figure~\ref{fig:fn}} presents the temporal accuracy comparison between MPDD and APDDM across multiple datasets. Each row corresponds to a dataset and shows accuracy over evaluation chunks (approximately 250 samples per chunk). The shaded green regions (AG) represent the accuracy gain achieved by APDDM over MPDD, illustrating the benefit of adaptive window adjustment under changing drift conditions. As shown in \textit{Fig.~\ref{fig:fn}(a)}, the HAR dataset highlights the advantage of adaptive monitoring under sudden and recurring drift. During major drift events, MPDD experiences substantial accuracy degradation, dropping from near-perfect performance to approximately 0.20, whereas APDDM maintains accuracy around 0.75 and recovers more rapidly. The corresponding AG regions indicate substantial accuracy improvements during drift intervals. The baseline methods exhibit larger fluctuations and less consistent behaviour across the stream. The Electricity results in \textit{Fig.~\ref{fig:fn}(b)} demonstrate the benefits of adaptation under mixed sudden and incremental drift. While MPDD accuracy falls to approximately 0.2 and later approaches 0.0 during major drift transitions, APDDM maintains higher accuracy and returns to stable performance more quickly, resulting in larger AG regions during the most severe drift periods. For Weather, \textit{Fig.~\ref{fig:fn}(c)} presents a smoother but under-sensitive drift pattern. MPDD accuracy drops to approximately 0.15 during major drift periods, whereas APDDM limits the degradation to around 0.4--0.6 and recovers earlier. Although the performance gap is smaller than in HAR and Electricity, the adaptive strategy consistently reduces both the magnitude and duration of accuracy loss, resulting in a smaller but consistent accuracy gain. The Airline dataset, shown in \textit{Fig.~\ref{fig:fn}(d)}, contains stronger sudden-drift behaviour and therefore represents a challenging setting for fixed monitoring. MPDD experiences severe performance drops, reaching nearly 0.0 during major drift events, while APDDM maintains near-perfect accuracy throughout most of the stream. Accordingly, the AG regions are most prominent during abrupt drift events, demonstrating the effectiveness of adaptive monitoring under severe drift conditions. Finally, \textit{Fig.~\ref{fig:fn}(e)} highlights the recurring and incremental drift characteristics of the Poker stream. MPDD exhibits several accuracy fluctuations throughout the stream, with noticeable drops to approximately 0.65 during drift intervals before recovering. In contrast, APDDM maintains more stable performance, remaining above approximately 0.85 for most of the evaluation period and recovering faster following drift events.

We further analyze the number of false negatives across different drift-detection techniques to evaluate their ability to identify real performance-drift events. Fig.~\ref{fig:false_negative} presents the comparison across datasets, where lower values indicate fewer missed drift detections. Overall, APDDM and MPDD achieve relatively low false-negative counts across all datasets. Although data-distribution-based techniques such as ADWIN, MMD, and D3 show lower false-negative counts for some datasets, particularly HAR and Electricity, these methods also produce significantly higher false-positive rates. Similarly, STUDD and SCSD achieve low false-negative counts in a few cases but exhibit substantially larger values on Weather and Poker, reducing their overall effectiveness. In contrast, MPDD and APDDM provide more consistent performance across different datasets. APDDM achieves the lowest or near-lowest false-negative counts in most cases, demonstrating that adaptive monitoring can reduce missed drift events while maintaining a better balance between false positives and false negatives. These results highlight the effectiveness of the proposed approaches for reliable MLaaS performance-drift detection in dynamic IoT environments.

\begin{table}[t]
\centering

\caption{Detection-quality comparison across drift detection techniques.}
\label{tab:detection_quality_all_methods}
\scriptsize
\renewcommand{\arraystretch}{1.08}
\setlength{\tabcolsep}{2.2pt}
\resizebox{\columnwidth}{!}{%
\begin{tabular}{llccccc}
\hline
\textbf{Data} & \textbf{Method} & \textbf{Acc.}$\uparrow$ & \textbf{Acc. Gain}$\uparrow$ & \textbf{MDR}$\downarrow$ & \textbf{FP}$\downarrow$ & \textbf{FN}$\downarrow$ \\
\hline
HAR & ADWIN & 0.62 & 30.42 & 0.56 & 190 & 270 \\
 & MMD & 0.52 & 40.00 & 0.66 & 260 & 315 \\
 & D3 & 0.30 & 62.50 & 0.73 & 495 & 350 \\
 & STUDD & 0.69 & 23.23 & 0.96 & 30 & 1465 \\
 & SCSD & 0.65 & 26.98 & 0.02 & 1650 & 25 \\
 & MPDD & 0.91 & 1.25 & 0.29 & 0 & 440 \\
 & \textbf{APDDM} & \textbf{0.92} & -- & 0.25 & 0 & 380 \\
\hline
Air & ADWIN & 0.65 & 31.00 & 0.44 & 35 & 35 \\
 & MMD & 0.65 & 31.00 & 0.44 & 35 & 35 \\
 & D3 & 0.60 & 36.00 & 1.00 & 0 & 80 \\
 & STUDD & 0.68 & 27.87 & 1.00 & 0 & 255 \\
 & SCSD & 0.42 & 54.12 & 0.00 & 465 & 0 \\
 & MPDD & 0.90 & 6.00 & 0.31 & 0 & 80 \\
 & \textbf{APDDM} & \textbf{0.96} & -- & 0.18 & 0 & 32 \\
\hline
Elec. & ADWIN & 0.77 & 15.35 & 0.31 & 85 & 125 \\
 & MMD & 0.69 & 23.05 & 0.47 & 90 & 190 \\
 & D3 & 0.72 & 20.30 & 0.19 & 180 & 75 \\
 & STUDD & 0.66 & 26.21 & 0.95 & 45 & 1185 \\
 & SCSD & 0.56 & 36.00 & 0.00 & 1585 & 0 \\
 & MPDD & 0.90 & 2.34 & 0.29 & 10 & 355 \\
 & \textbf{APDDM} & \textbf{0.92} & -- & 0.24 & 0 & 280 \\
\hline
Weath. & ADWIN & 0.50 & 39.19 & 0.62 & 305 & 655 \\
 & MMD & 0.54 & 35.82 & 0.60 & 260 & 635 \\
 & D3 & 0.57 & 32.72 & 0.58 & 225 & 610 \\
 & STUDD & 0.61 & 28.92 & 0.95 & 235 & 2810 \\
 & SCSD & 0.69 & 20.30 & 0.80 & 35 & 2345 \\
 & MPDD & 0.89 & 0.79 & 0.30 & 5 & 870 \\
 & \textbf{APDDM} & \textbf{0.90} & -- & 0.27 & 20 & 794 \\
\hline
Poker & ADWIN & 0.54 & 43.25 & 0.61 & 430 & 485 \\
 & MMD & 0.54 & 43.25 & 0.62 & 420 & 495 \\
 & D3 & 0.53 & 45.00 & 0.72 & 375 & 575 \\
 & STUDD & 0.68 & 29.69 & 1.00 & 30 & 2545 \\
 & SCSD & 0.70 & 27.50 & 0.94 & 5 & 2395 \\
 & MPDD & 0.88 & 9.81 & 0.25 & 345 & 640 \\
 & \textbf{APDDM} & \textbf{0.98} & -- & 0.06 & 40 & 160 \\
\hline
\end{tabular}%
}

\end{table}

\textit{Table~\ref{tab:detection_quality_all_methods}} presents the comparative performance of the proposed APDDM framework against the fixed-interval MPDD baseline and five representative drift detection techniques, namely ADWIN, MMD, D3, STUDD, and SCSD, across five datasets. The comparison is conducted using Accuracy$\uparrow$, Accuracy Gain$\uparrow$, Miss Detection Ratio (MDR)$\downarrow$, and False Negatives (FN)$\downarrow$. Accuracy reflects the overall correctness of drift detection decisions, while Accuracy Gain quantifies the relative improvement achieved by APDDM over competing approaches. MDR and FN are used to evaluate the ability of each method to identify drift events without missing significant changes in the data stream. Lower MDR and FN values indicate better drift detection capability and improved responsiveness to evolving data distributions. Overall, APDDM consistently achieves the highest detection accuracy across all datasets while simultaneously reducing the number of missed drift events. For the \textit{HAR} dataset, APDDM achieves the highest accuracy of 0.92, representing an improvement of 1.25\% over the MPDD baseline. Furthermore, APDDM reduces the number of false negatives from 440 under MPDD to 380, corresponding to a reduction of approximately 13.6\%, while also lowering the MDR from 0.28 to 0.25. These results indicate that adaptive monitoring improves drift-detection accuracy while reducing missed drift events. A similar trend is observed in the \textit{Airline} dataset, where APDDM achieves an accuracy of 0.96 compared to 0.90 for MPDD, representing a 6.0\% improvement. In addition, the number of false negatives is reduced from 80 to 32, corresponding to a 60.0\% reduction, while the MDR decreases from 0.31 to 0.17. This demonstrates the effectiveness of adaptive monitoring in identifying evolving drift patterns and minimizing missed detections. For the \textit{Electricity} dataset, APDDM achieves the highest accuracy of 0.92 and reduces false negatives from 355 under MPDD to 280, corresponding to a reduction of approximately 21.1\%. Similarly, the MDR decreases from 0.285 to 0.23. Although D3 achieves a lower MDR of 0.18 with only 75 false negatives, its overall detection accuracy remains substantially lower at 0.72, indicating a less balanced trade-off between drift coverage and overall detection quality. In the \textit{Weather} dataset, APDDM again provides the highest accuracy of 0.89 while reducing false negatives from 870 to 794 and lowering the MDR from 0.29 to 0.26 compared with MPDD. The performance gap is even more pronounced when compared with traditional drift-detection techniques, which exhibit substantially higher MDR values and considerably larger numbers of missed drift events. Similarly, for the \textit{Poker} dataset, APDDM achieves the highest accuracy of 0.97, representing a 9.81\% improvement over MPDD. Moreover, the number of false negatives is reduced dramatically from 640 under MPDD to only 160, corresponding to a 75.0\% reduction. This improvement is accompanied by a substantial decrease in MDR from 0.25 to 0.06, demonstrating the effectiveness of adaptive monitoring in reducing missed drift detections. In contrast, methods such as STUDD and SCSD exhibit MDR values close to 1.0 and very large false-negative counts, indicating that a substantial proportion of drift events remain undetected. Across all datasets, the results demonstrate that APDDM consistently provides a more favourable balance between detection accuracy and drift coverage.

\subsection{Experiment 3: Comprehensive Evaluation of the MLaaS Extraction Model}
The MLaaS extraction model is key in the proposed MPDD framework to approximate the behaviour of the original black-box MLaaS service. The effectiveness of the framework depends on the quality of this approximation. To evaluate this, we first analyze the fidelity and prediction behavior analysis of the MLaaS extracted model. Next, we then investigate the impact of MLaaS extraction model error on the MPDD and APDDM techniques performance.
\subsubsection{Fidelity and Predictive Behavior Analysis of the MLaaS extraction Model} 
\begin{figure*}[t]
\centering
\subfloat[]{
    \includegraphics[width=\columnwidth]{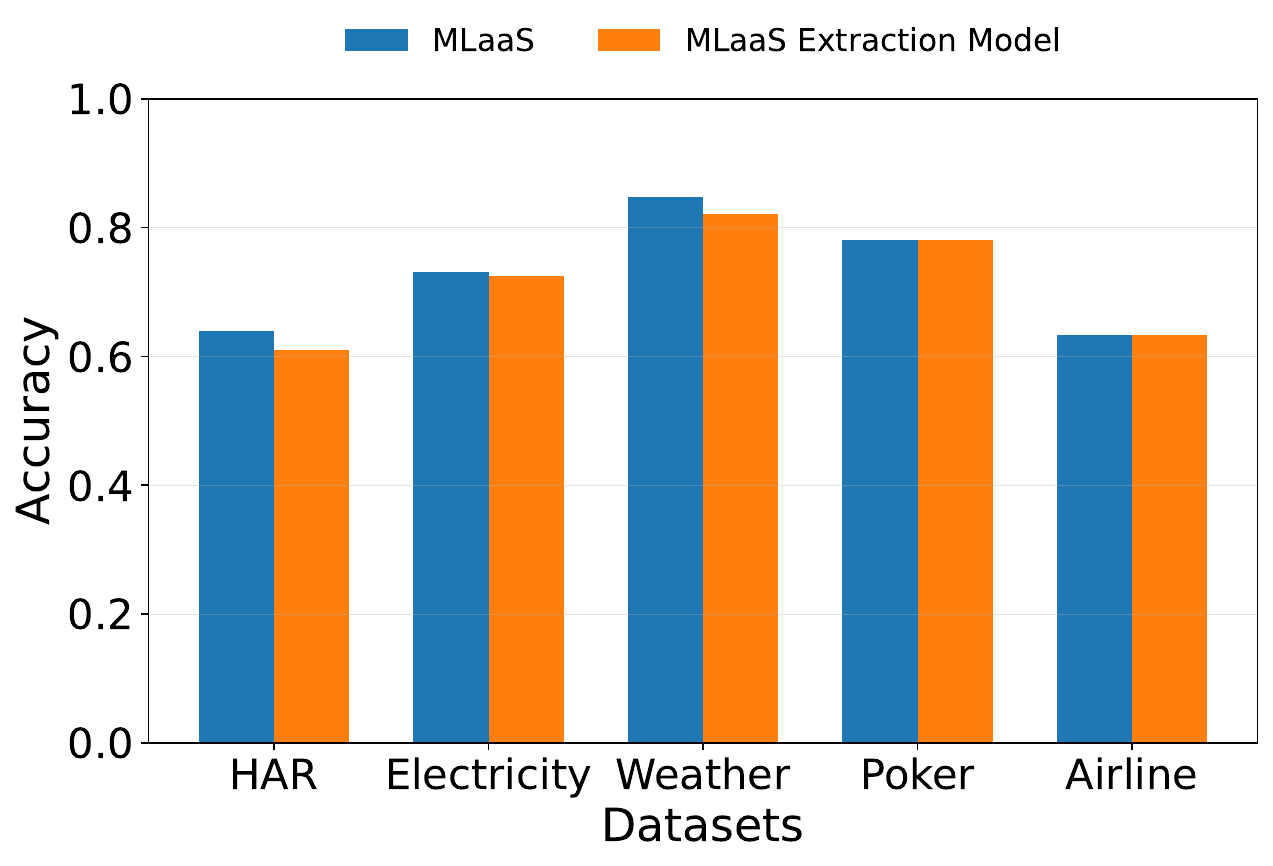}
    \label{fig:surrogate_accuracy}
}
\subfloat[]{
    \includegraphics[width=\columnwidth]{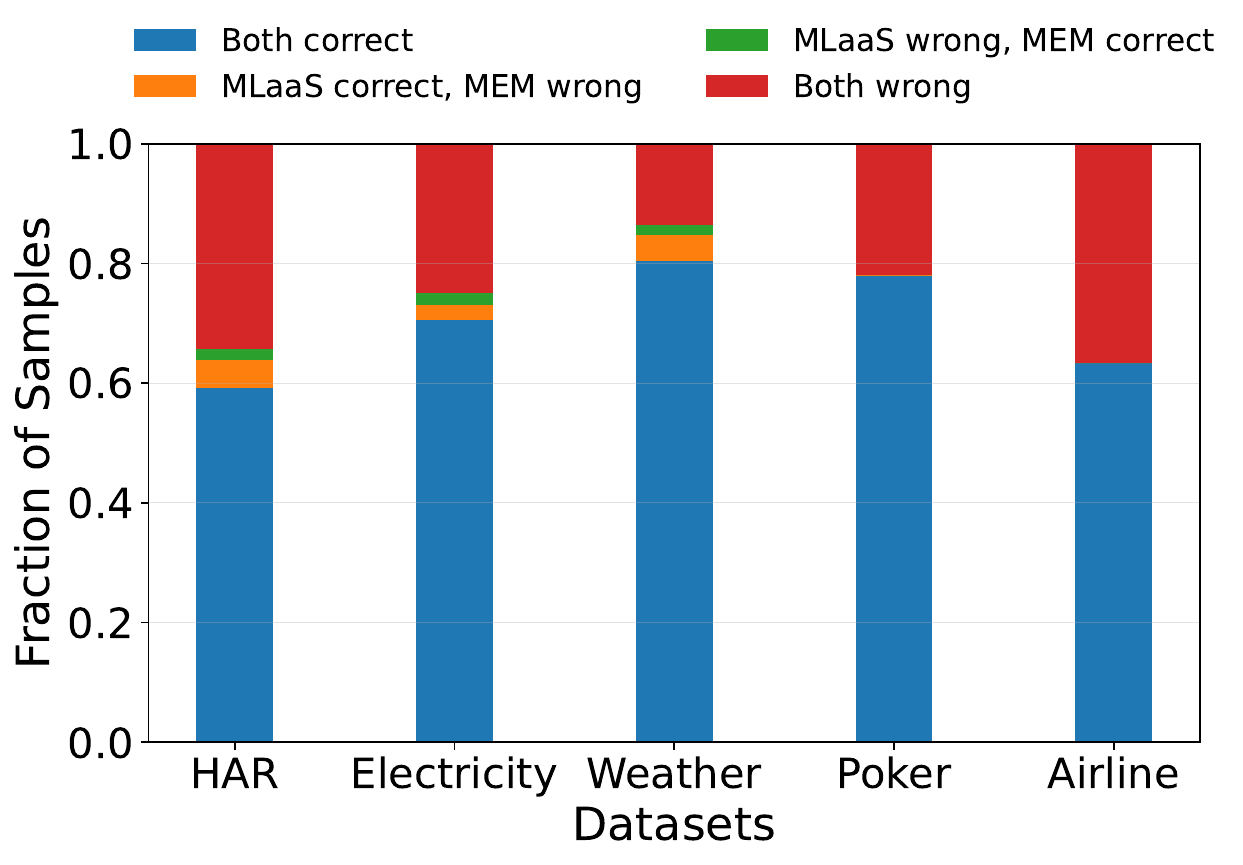}
    \label{fig:surrogate_agreement}
}
\caption{\normalsize Fidelity and predictive behavior analysis of the MLaaS extraction model across datasets: (a) predictive accuracy comparison, and (b) predictive behavior analysis}
\label{fig:surrogate_fidelity}
\end{figure*}

In this experiment, we evaluate the fidelity and predictive behavior of the proposed MLaaS extraction model, which aims to approximate the decision patterns of black-box MLaaS services. Fidelity analysis is a standard evaluation strategy in model extraction research. It measures the agreement between the predictions of the extracted model and those of the original inaccessible MLaaS service, rather than the accuracy with respect to the ground-truth labels ~\cite{jagielski2020high}. Figure~\ref{fig:surrogate_fidelity}(a) compares the predictive accuracy of both the original MLaaS service and the extraction model against the ground truth across all datasets after performance drift has occurred. As expected, the predictive accuracy of the original MLaaS service decreases under drifted conditions. The extraction model exhibits nearly the same degradation trend, closely tracking the predictive behaviour of the original MLaaS service. The deviation between the MLaaS service and the extraction model remains small across all datasets, at approximately 3\% for HAR, 1\% for Electricity, 3\% for Weather, and nearly identical performance for Poker and Airline. Consequently, the extraction model achieves an average fidelity of approximately \textit{95.2\%}, indicating a high level of agreement with the predictions of the original MLaaS service. Specifically, the extraction model achieves predictive accuracies against the ground truth of approximately 0.61 on HAR, 0.72 on Electricity, 0.82 on Weather, 0.78 on Poker, and 0.63 on Airline, closely matching the corresponding accuracies of the original MLaaS services under the same drift conditions. These results demonstrate that the extraction model effectively captures original MLaaS service. To further investigate behavioral consistency, Fig.~\ref{fig:surrogate_fidelity}(b) presents a predictive outcome decomposition that categorizes predictions into four cases: both MLaaS and extraction models correct, MLaaS correct but extraction incorrect, extraction correct but MLaaS incorrect, and both models incorrect. This analysis provides a more detailed understanding of the relationship between the extracted model and the original MLaaS service beyond overall accuracy. 

Across datasets, the dominant portion corresponds to the both-correct category, reaching approximately 0.80 for Weather, 0.78 for Poker, 0.70 for Electricity, 0.63 for Airline, and 0.59 for HAR, indicating substantial agreement between the models. The disagreement regions remain relatively small, with the MLaaS-correct/extraction-wrong and MLaaS-wrong/extraction-correct categories generally below 5\% across datasets. The both-wrong category is primarily influenced by the inherent difficulty of each dataset. However, the small disagreement fractions indicate that the extraction model closely follows the predictive behavior of the original MLaaS service.

\subsubsection{Impact of MLaaS Extraction Model Errors on MPDD and APDDM Performance}
\begin{figure}[t]
\centering
\includegraphics[width=\columnwidth]{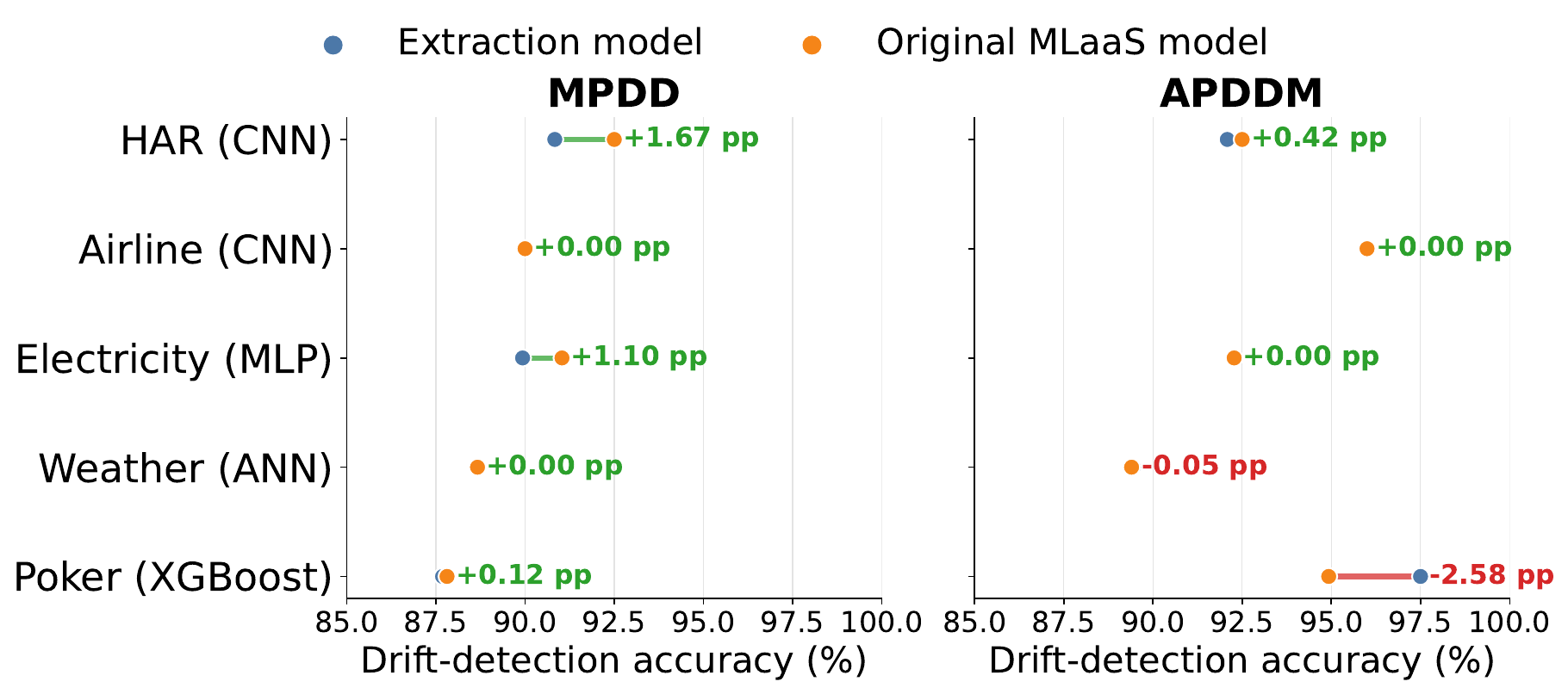}
\caption{\normalsize Comparison of MPDD and APDDM accuracy when using the MLaaS extracted model and the original MLaaS}
\label{fig:mpdd_surrogate_comparison}
\end{figure}

A key concern in black-box MLaaS monitoring is that the extracted model may not perfectly reproduce the decision behaviour of the original service, particularly when the underlying model is a complex non-linear model. Such approximation errors may occur in disagreement regions and could potentially affect downstream drift-detection decisions. Therefore, we quantify how much MLaaS extraction-model errors propagate to the final drift-detection outcome. Figure~\ref{fig:mpdd_surrogate_comparison} compares the drift-detection accuracy obtained using the MLaaS extracted model and the original black-box MLaaS model for both MPDD and APDDM. For MPDD, the results show only small differences between the two models, with accuracy differences ranging from \textit{0.00 to 1.67 percentage} points across datasets. No difference is observed for Airline and Weather, while HAR, Electricity, and Poker exhibit only minor variations. For APDDM, the results remain highly consistent with those obtained using the original MLaaS model. The accuracy remains unchanged for Airline, Electricity, and Weather, with only a small difference for HAR and a moderate decrease for Poker. Overall, APDDM is even less sensitive to extraction-model approximation errors, with deviations ranging from \textit{-2.58 to +0.42 percentage points}, and in most cases showing no measurable difference. This robustness stems from APDDM's adaptive parameter-selection mechanism, which adjusts the monitoring interval based on recent drift-score trends rather than relying on the output of any single extraction-model prediction. Consequently, small approximation errors have only a limited influence on the final drift-detection decision. These results indicate that MLaaS extraction-model errors have limited impact on the final MPDD and APDDM decisions across the evaluated scenarios. Hence, the extracted model provides a reliable approximation of the original MLaaS service for performance-drift detection.

%Our framework enables consumers to identify potential drifts in MLaaS, allowing them to notify providers, opt for alternative services, or perform service composition. Since some drift points in the experiments were generated randomly, the results may be sensitive to their placement. To mitigate this, we repeated the random generation multiple times and reported aggregated metrics to ensure robustness. Moreover, model replacement or service switching is a key business decision for clients. While the proposed framework is effective in detecting performance drift, it currently does not distinguish between different drift types that may arise in dynamic IoT environments. Clients should also understand the underlying causes of detected drift to make informed and sustainable adaptation decisions. Extending this framework to incorporate feature-level analysis and adaptive models capable of handling diverse drift patterns would further enhance its practicality and robustness.
%Although the proposed framework employs a lightweight extraction model for efficient runtime monitoring, the fidelity of behavioral approximation may vary for highly complex black-box MLaaS models with non-linear decision boundaries.

\section{Discussions and Implications}
Our framework enables clients to identify potential drifts in MLaaS, allowing them to notify providers, opt for alternative services, or perform service composition. Since some drift points in the experiments were generated randomly, the results may be sensitive to their placement. To mitigate this, we repeated the random generation multiple times and reported aggregated metrics to ensure robustness. In this context, the proposed approach functions as an effective service management tool for MLaaS consumers by enabling proactive drift awareness, informed provider notification, and timely service switching or composition decisions. This capability is especially important in cost-sensitive and safety-critical domains, such as smart healthcare, industrial IoT, and intelligent transportation, where undetected performance drift can lead to incorrect decisions, operational risks, and financial losses.

\section{Concluding Remarks}
In this paper, we proposed an MLaaS Performance Drift Detection (MPDD) framework for IoT environments where ground-truth labels and internal model parameters are inaccessible. To address this limitation, we designed an MLaaS extraction model that captures observable behavioral patterns of black-box MLaaS services, enabling reliable monitoring of feature preference variations. We further introduced a Fréchet-based drift measurement mechanism to quantify distributional changes in evolving data streams and developed an MLaaS-aware Drift Exposure Score within MPDD to predict performance drift. Extensive experiments across multiple real-world datasets demonstrate that the framework improves drift detection effectiveness, achieving approximately 22–25\% accuracy improvement over representative baseline drift detection methods while maintaining stability under diverse drift patterns. Additionally, the adaptive-interval APDDM dynamically adjusts monitoring frequency based on drift intensity, yielding around 4\% accuracy improvement and approximately 9\% reduction in miss detection rate compared to fixed-interval monitoring, thereby enhancing detection reliability and timeliness in non-stationary IoT streams. Overall, the framework provides a robust and timely solution for continuous MLaaS performance monitoring without requiring manual drift inspection or continuous ground-truth collection.\\

\bibliographystyle{unsrt}
\bibliography{references}

\appendices
\section{Proof of Lemmas}\label{ap1}
\subsection*{\textbf{Proof of Lemma 1}}
Let $M$ be the black-box MLaaS and $M'$ be the extraction model trained to approximate $M$ from query logs. 
Assume the bounded fidelity condition holds:
\[
\mathbb{E}\big[\,|M(X)-M'(X)|\,\big]\le \epsilon .
\]

Fix a feature $x_i$. In Algorithm~1, feature importance is defined operationally (i.e., from input--output behavior) as the average change in the model output under a controlled perturbation of magnitude $\delta$ applied to $x_i$. Let $X_{+i}$ denote the perturbed version of $X$ along $x_i$. Consider the (per-sample) importance contributions
\[
\begin{aligned}
\Delta_i^M(X)
&= \big|M(X_{+i})-M(X)\big|, \\
\Delta_i^{M'}(X)
&= \big|M'(X_{+i})-M'(X)\big|.
\end{aligned}
\]

By the triangle inequality,
\[
\begin{aligned}
\big|\Delta_i^{M'}(X)-\Delta_i^{M}(X)\big|
&\le 
\big|M'(X_{+i})-M(X_{+i})\big| \\
&\quad + \big|M'(X)-M(X)\big|.
\end{aligned}
\]
Taking expectation and using the fidelity assumption on both $X$ and $X_{+i}$ gives
\[
\mathbb{E}\big[\,|\Delta_i^{M'}(X)-\Delta_i^{M}(X)|\,\big]\;\le\; 2\epsilon .
\]
Averaging over the samples used by Algorithm~1 yields the same bound for the estimated importance of feature $x_i$:
\[
\mathbb{E}\big[\,|I^{M'}(x_i)-I^{M}(x_i)|\,\big]\;\le\; 2\epsilon .
\]
Therefore, the extracted importance vector $\mathbf{I}^{M'}$ deviates from the behavioural importance vector $\mathbf{I}^{M}$ by a bounded amount. Since the measured importance is induced by perturbations of magnitude $\delta$, the estimator sensitivity is governed by $\delta$, yielding
\[
\|\mathbf{I}^{M'}-\mathbf{I}^{M}\|\;\le\; f(\epsilon,\delta),
\]
for some bounded function $f(\epsilon,\delta)$ that is monotone in $\epsilon$ and dependent on the perturbation scale $\delta$ and the chosen norm. Hence, under bounded fidelity, the extraction model preserves the feature-importance behaviour of the black-box MLaaS up to a bounded deviation, establishing Lemma~1.
\hfill\ensuremath{\square}
\vspace{-2mm}
\begin{comment}

\end{comment}
\subsection*{\textbf{Proof of Lemma 2}}
Recall that FDDS between the baseline window $b$ and current window $w$ is defined as in Eq.~(5) of the paper:
\[
\mathrm{FDDS}(b,w)
= \underbrace{\|\mu_b-\mu_w\|_2^2}_{mean shift}
+ \underbrace{\mathrm{Tr}\!\Big(\Sigma_b+\Sigma_w-2\sqrt{\Sigma_b\Sigma_w}\Big)}_{covariance shift}.
\]
\smallskip
\noindent\textbf{(i) Lower bounded by the mean shift.} Let $\mathcal{C}(\Sigma_b,\Sigma_w)$ denote the covariance shift term. Since $\Sigma_b$ and $\Sigma_w$ are symmetric positive semi-definite, the trace-based covariance discrepancy is non-negative and cannot reduce the FDDS below the mean shift component. Hence,
\[
\mathrm{FDDS}(b,w)=\|\mu_b-\mu_w\|_2^2+\mathcal{C}(\Sigma_b,\Sigma_w)
\;\ge\;\|\mu_b-\mu_w\|_2^2,
\]
which shows that FDDS is lower bounded by the mean shift.\\
\smallskip
\begin{comment}

\end{comment}
\noindent\textbf{(ii) Under significant covariance change}
Consider that the current window undergoes a substantial  covariance drift case where the current covariance differs substantially from the baseline, i.e., $\|\Sigma_w-\Sigma_b\|$ is large. Then the product $\Sigma_b\Sigma_w$ changes accordingly, and the square-root interaction term $\sqrt{\Sigma_b\Sigma_w}$ no longer aligns with $\Sigma_b$.
As a result, the gap
\[
\Sigma_b+\Sigma_w-2\sqrt{\Sigma_b\Sigma_w}
\]
becomes larger in the trace sense, so $\mathcal{C}(\Sigma_b,\Sigma_w)$ increases, which directly increases $\mathrm{FDDS}(b,w)$ even if $\mu_b\approx\mu_w$.\\\smallskip
\noindent\textbf{(iii) Under minor distributional change}
Let us consider the case where the distributional change between the baseline and current windows is minor, meaning $\|\mu_b-\mu_w\|$ is small and $\Sigma_w\approx\Sigma_b$, then the mean shift term is small and
$\sqrt{\Sigma_b\Sigma_w}\approx\sqrt{\Sigma_b^2}=\Sigma_b$, making $\mathcal{C}(\Sigma_b,\Sigma_w)\approx 0$.
Therefore, $\mathrm{FDDS}(b,w)$ stays small under minor fluctuations. Combining (i)–(iv), FDDS is lower bounded by the mean shift and increases with either substantial mean drift or substantial covariance drift, while remaining low under minor changes. This proves Lemma~2.
\hfill\ensuremath{\square}
\vspace{-2mm}
\subsection*{\textbf{Proof of Lemma 3}}
\textit{Proof.}
As defined in Eq.~(6), Algorithm~2  flags drift when the \emph{MDES} score exceeds the threshold $\theta_m$, i.e.,
\[
\mathrm{MDES} =
\begin{cases}
1, & \text{if }\displaystyle\sum_{i=1}^{d} I(x_i)\cdot \mathrm{FDDS}(b_i,w_i) > \theta_m,\\
0, & \text{otherwise.}
\end{cases}
\]
Consider the MDES defined in Eq.~(6), where the cumulative exposure is given by $\sum_{i=1}^{d} I(x_i)\cdot \mathrm{FDDS}(b_i,w_i)$, which jointly captures the feature importance vector $\mathbf{I}$ and the feature-wise statistical shift between $b$ and $w$. If the statistical drift is predominantly concentrated on features with high importance weights, then the corresponding terms $I(x_i)\cdot \mathrm{FDDS}(b_i,w_i)$ become large, leading to a significant increase in the cumulative weighted exposure. Consequently, the exposure is more likely to exceed the threshold $\theta_m$, which activates the condition $\mathrm{MDES}=1$ and indicates that the drift is aligned with influential features, thereby characterizing it as real drift. In contrast, if the statistical shift mainly occurs on features with low importance weights, the same $\mathrm{FDDS}(b_i,w_i)$ contributions are attenuated by small $I(x_i)$ values, keeping the cumulative exposure bounded and unlikely to surpass $\theta_m$. Under this condition, $\mathrm{MDES}=0$, and the drift is interpreted as pseudo-drift. Therefore, the MLaaS Drift Exposure Score distinguishes real drift from pseudo-drift based on the alignment between drift magnitude and feature importance. This proves Lemma~3.
\hfill\ensuremath{\square}

\section{Drift Generation}\label{app2}

\begin{table}[H]
\renewcommand{\thetable}{6}
\centering
\caption{Drift generation statistics across datasets.}
\label{tab:drift_generation_statistics}
\scriptsize
\renewcommand{\arraystretch}{1.08}
\setlength{\tabcolsep}{2.2pt}
\resizebox{\columnwidth}{!}{%
\begin{tabular}{lrrrrrr}
\hline
\textbf{Dataset} &
\textbf{No Drift} &
\textbf{Sudden} &
\textbf{Incremental} &
\textbf{Gradual} &
\textbf{Recurrent} &
\textbf{Total Drift} \\
\hline

HAR &
60,000 &
24,000 &
14,400 &
14,000 &
24,000 &
76,400 \\

Airline &
9,989 &
3,996 &
2,398 &
2,332 &
3,998 &
12,724 \\

Electricity &
45,312 &
20,391 &
12,235 &
11,329 &
18,126 &
62,081 \\

Weather &
96,453 &
53,050 &
31,830 &
24,109 &
38,583 &
147,572 \\

Poker &
100,000 &
40,000 &
24,000 &
23,336 &
40,000 &
127,336 \\

\hline
\end{tabular}%
}
\end{table}
\begin{figure}[!h]
\renewcommand{\thefigure}{8}
\centering
\subfloat[]{
  \includegraphics[width=\columnwidth,height=5.2cm,keepaspectratio]{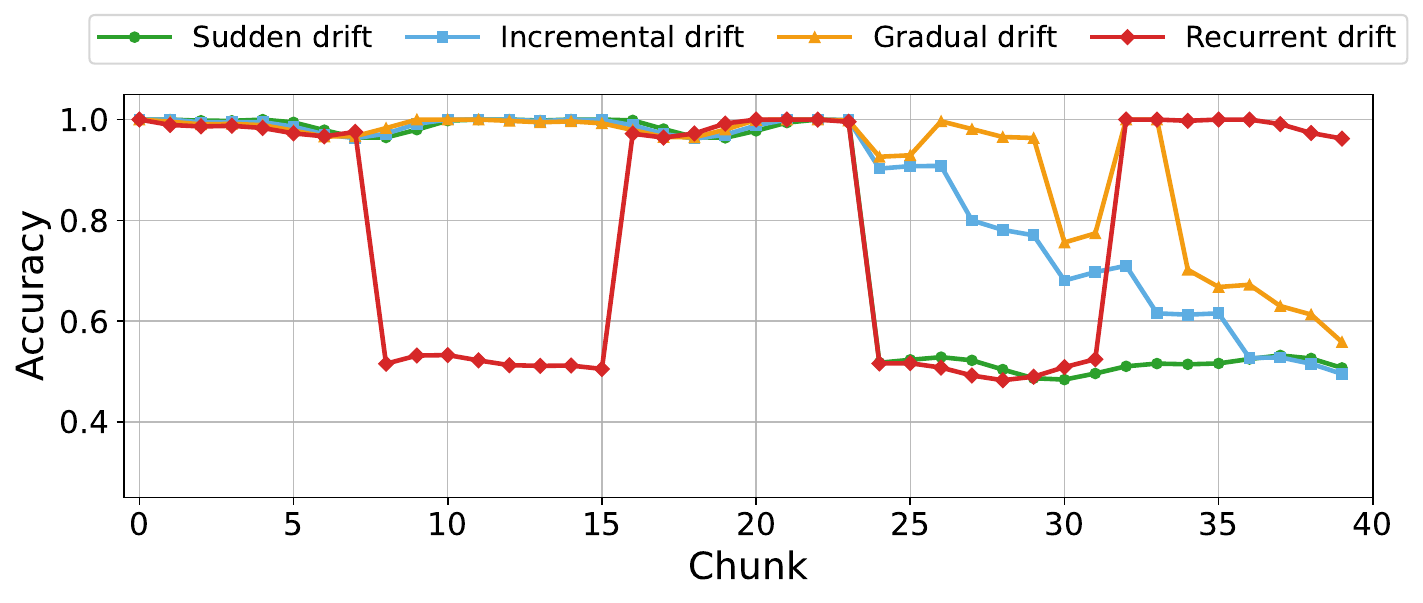}
  \label{fig:gen-a}
}
\par\vspace{1mm}

\subfloat[]{
\includegraphics[width=\columnwidth,height=5.2cm,keepaspectratio]{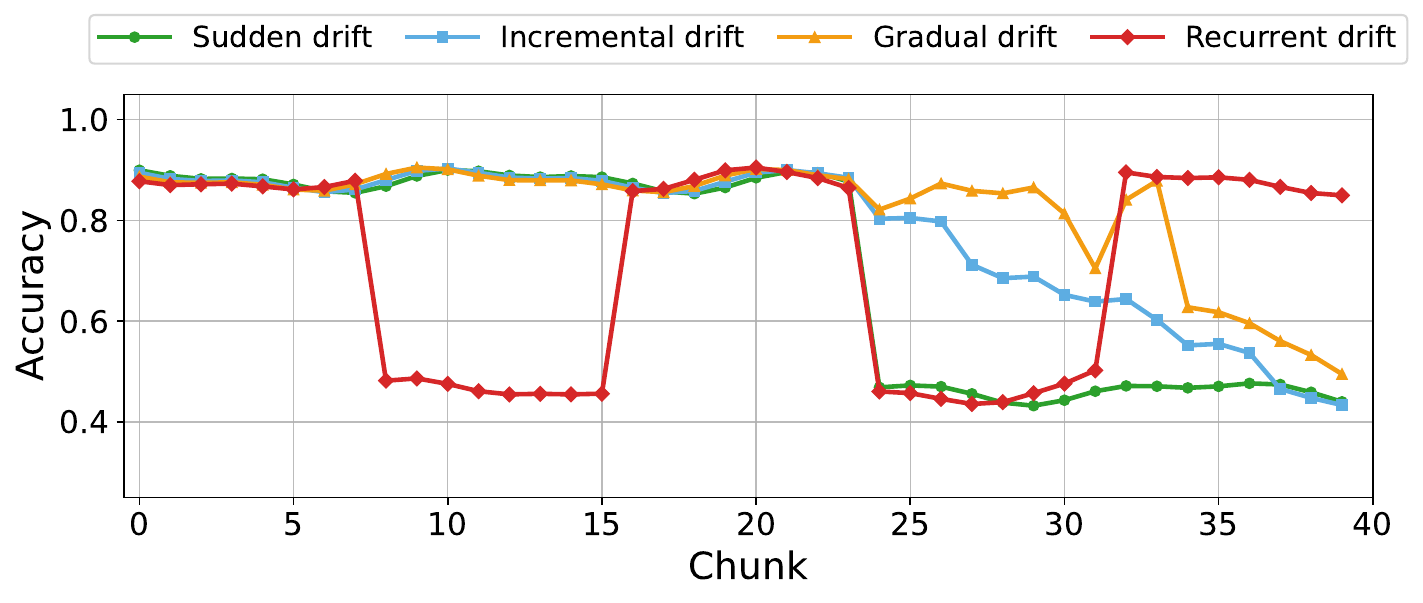}
  \label{fig:gen-b}
}
\par\vspace{1mm}

\subfloat[]{
  \includegraphics[width=\columnwidth,height=5.2cm,keepaspectratio]{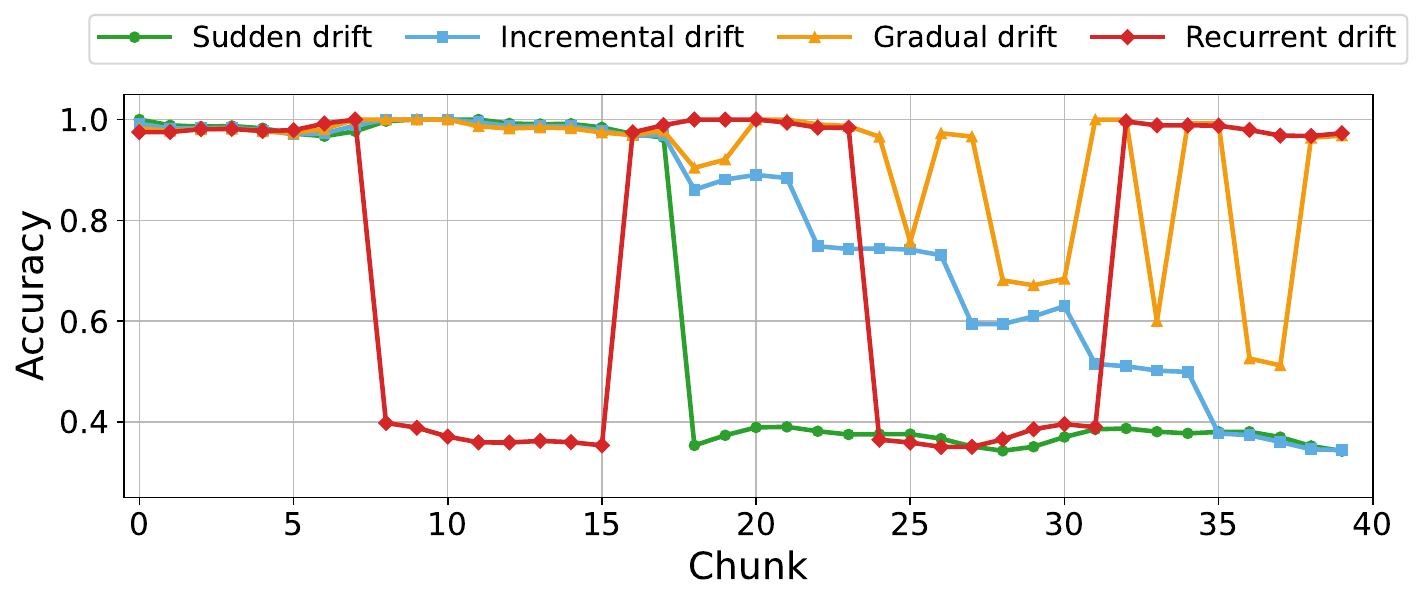}
  \label{fig:gen-c}
}
\par\vspace{1mm}

\subfloat[]{
  \includegraphics[width=\columnwidth,height=5.2cm,keepaspectratio]{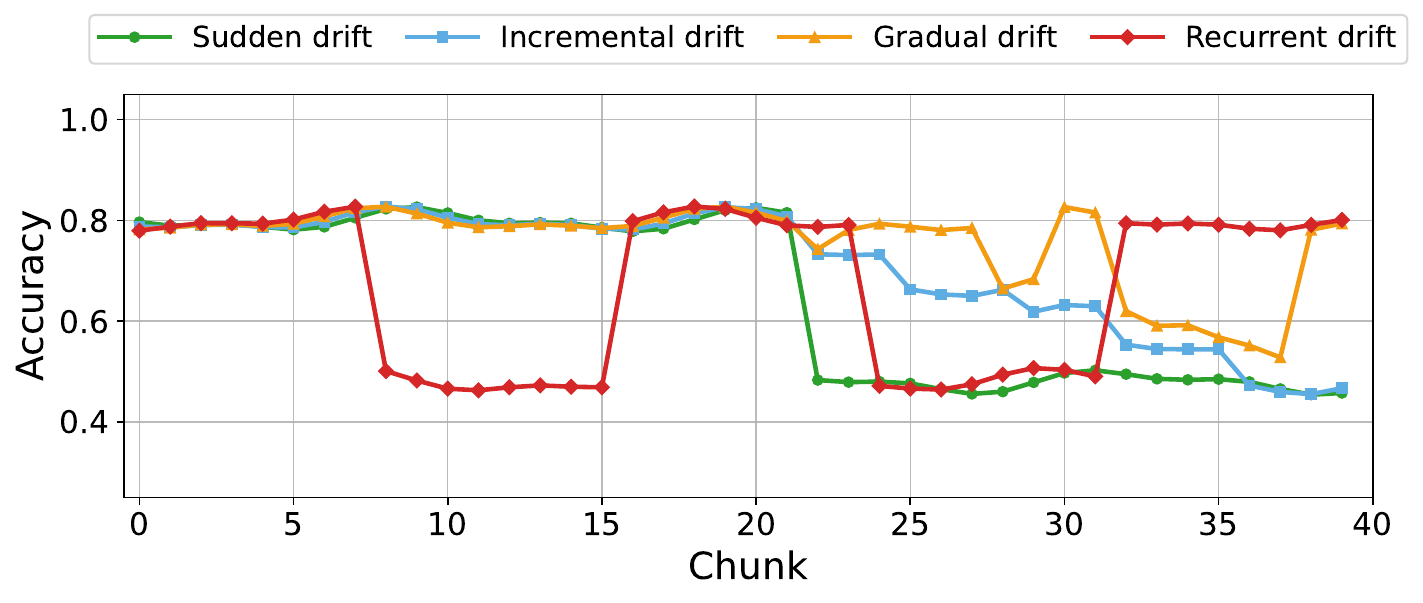}
  \label{fig:gen-d}
}
\par\vspace{1mm}

\subfloat[]{
  \includegraphics[width=\columnwidth,height=5.2cm,keepaspectratio]{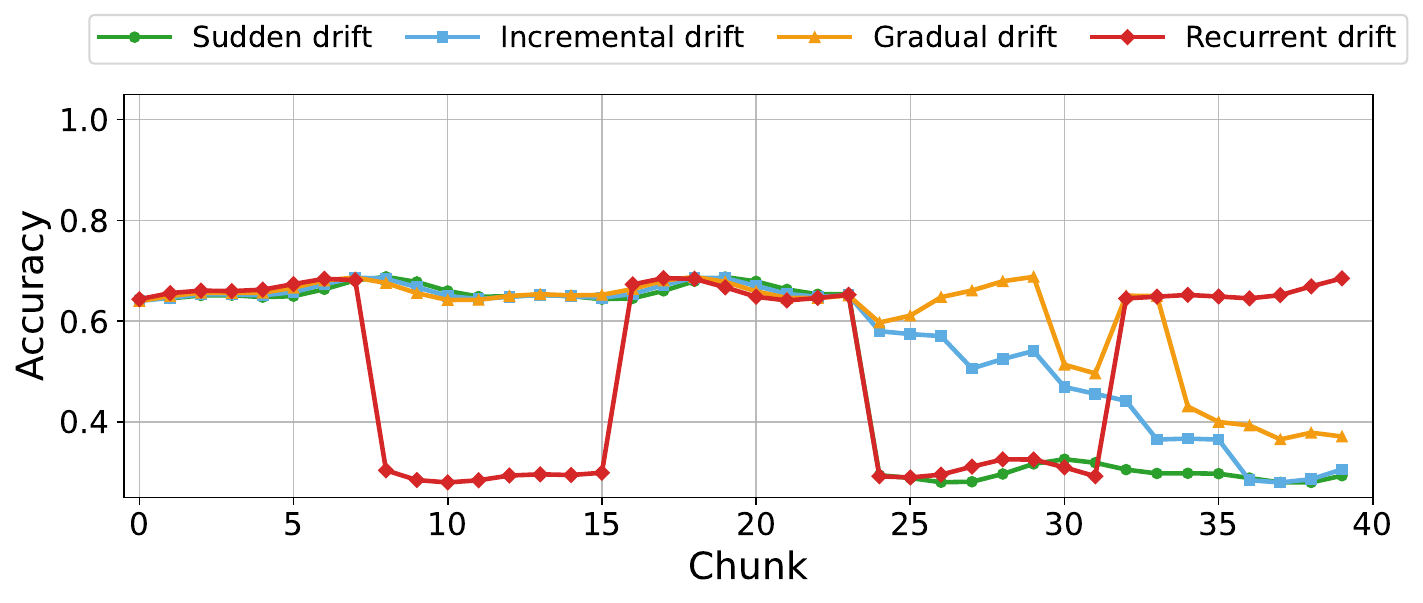}
  \label{fig:gen-e}
}
\caption{Drift generation across datasets: (A) HAR, (B) Airline, (C) Weather, (D) Electricity, and (E) Poker.}
\label{fig:motivation_generators}
\vspace{-3mm}
\end{figure}

\subsubsection{Drift Data Generation}
We use the SEA generator \cite{pedregosa2011scikit} to create drifted data streams across the datasets used in our experiments, modelling four common drift categories: sudden, incremental, gradual, and recurrent drift. For each dataset, four full-length drifted streams are generated, one for each drift type. Consequently, the combined generated stream contains four times the number of samples in the original dataset. Table~\ref{tab:drift_generation_statistics} summarizes the stream sizes and drift statistics across all datasets. Specifically, the generated streams contain 240,000 samples for HAR, 39,956 for Airline, 181,248 for Electricity, 385,812 for Weather, and 400,000 for Poker.  The generated drift regions comprise 76,400 samples for HAR, 12,724 for Airline, 62,081 for Electricity, 147,572 for Weather, and 127,336 for Poker. These drift samples are distributed across sudden, incremental, gradual, and recurrent drift scenarios. For each dataset, we generate one sudden drift event, one gradual drift event, one incremental drift event consisting of five progressively stronger stages, and two recurrent drift events, resulting in five drift episodes per dataset and 25 drift episodes across all datasets. Figure \ref{fig:motivation_generators} illustrates the drift behaviour across the generated streams and highlights how different drift types evolve over time. For evaluation, each stream is processed using chunks of 250 samples, where each chunk consists of five consecutive blocks of 50 samples. A chunk is labelled as a real-drift chunk when more than 50\% of its samples belong to a generated drift region. The visualizations presented in Appendix B show uniformly selected representative chunks from the complete streams and are intended to illustrate drift progression rather than indicate the total number of generated drift events.

\begin{IEEEbiography}
[{\includegraphics[width=1in,height=1.25in,clip,keepaspectratio]{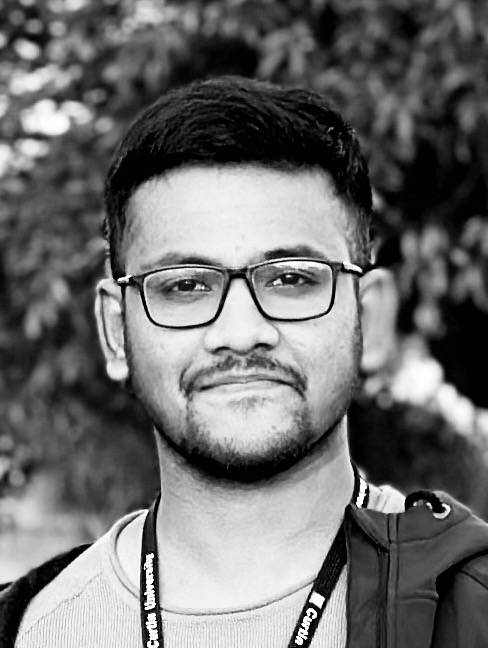}}]{Deepak Kanneganti} is an Associate Lecturer at Curtin University, Australia. He received his Master's degree in Predictive Analytics from Curtin University, Bentley, Perth, WA, Australia, and his Bachelor's degree in Electronics and Communication Engineering. He previously served as a Data Science Intern at the Pawsey Supercomputing Research Centre. His research interests include distributed machine learning, federated learning, and explainable artificial intelligence.
\end{IEEEbiography}
\vspace{-10mm}
\begin{IEEEbiography}
[{\includegraphics[width=1in,height=1.25in,clip,keepaspectratio]{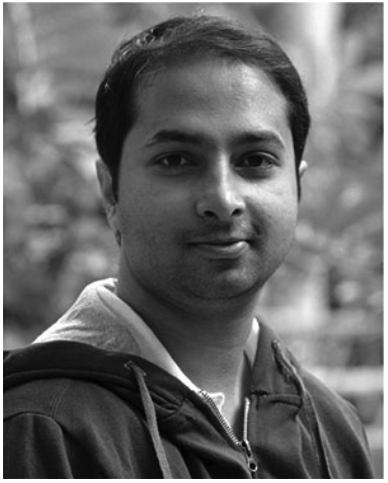}}]{Sajib Mistry} is an Associate Professor at Curtin University, Australia. He received his Ph.D. degree from RMIT University, Australia. Prior to joining Curtin University, he was a Postdoctoral Fellow with the School of Computer Science, University of Sydney, Australia. His research interests include edge and cloud computing, big data, and the Internet of Things. He has published articles in international journals and conferences, including IEEE Transactions on Services Computing, IEEE Transactions on Knowledge and Data Engineering, Communications of the ACM, ICSOC, WISE, and IEEE ICWS. He received the Best Paper Award at ICSOC 2016.
\end{IEEEbiography}
\vspace{-10mm}
\begin{IEEEbiography}
[{\includegraphics[width=1in,height=1.25in,clip,keepaspectratio]{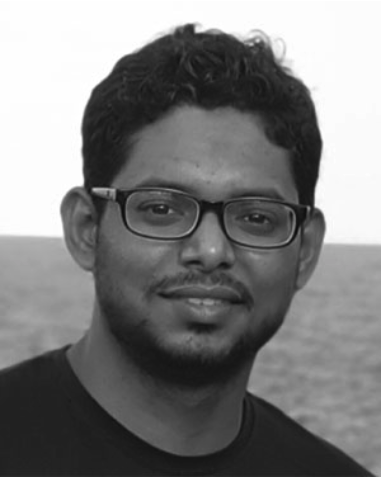}}]{Sheik Mohammad Mostakim Fattah} is a Lecturer at Curtin University, Australia. He received his Ph.D. degree in Computer Science from the University of Sydney, Australia, an M.Eng. degree in Computer and Information Communication Engineering from Hankuk University of Foreign Studies, South Korea, and a B.Sc. (Hons.) degree in Computer Science and Engineering from the University of Dhaka, Bangladesh. Prior to joining Curtin University, he held academic and research positions at the University of Adelaide, Torrens University, and the University of Sydney, as well as industry positions at the Korea Electronics Technology Institute and Monist IT Ltd. His research interests include cloud computing, services computing, edge computing, the Internet of Things, and Semantic Web technologies. His work has appeared in leading venues such as IEEE Transactions on Services Computing, ACM Transactions on the Web, ICSOC, and IEEE ICWS. He serves as a reviewer and program committee member for several international conferences and journals.
\end{IEEEbiography}
\vspace{-10mm}
\begin{IEEEbiography}
[{\includegraphics[width=1in,height=1.25in,clip,keepaspectratio]{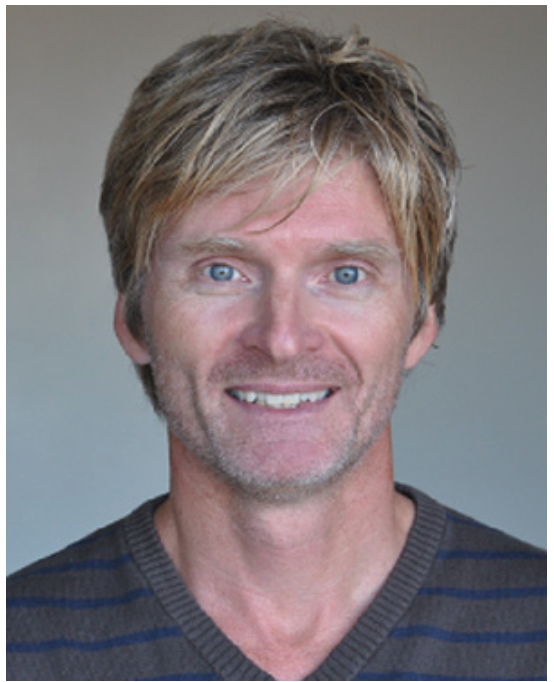}}]{Erik Elmroth} (Member, IEEE) is a Professor of Computing Science at Umeå University, Sweden. He has served as Head and Deputy Head of the Department of Computing Science for 13 years and as Deputy Director of the National Supercomputer Centre for another 13 years. He established Umeå University's research activities in distributed systems. His experience in management and executive groups of large-scale research initiatives includes the EUR 550 million Wallenberg AI, Autonomous Systems and Software Program and the strategic research area eSSENCE. He has developed two international research strategies for the Nordic Council of Ministers. His international experience includes one year at NERSC, Lawrence Berkeley National Laboratory, University of California, Berkeley, USA, and one semester at the Massachusetts Institute of Technology, Cambridge, MA, USA. He has served as a member of the Swedish Research Council's Committee for Research Infrastructure, Chair of its expert panel on eScience, and Chair of the Board of the Swedish National Infrastructure for Computing. He is a Lifetime Member of the Royal Swedish Academy of Engineering Sciences and has served as Vice Chair of its Division for Information Technology.
\end{IEEEbiography}
\vspace{-13mm}
\begin{IEEEbiography}
[{\includegraphics[width=1in,height=1.25in,clip,keepaspectratio]{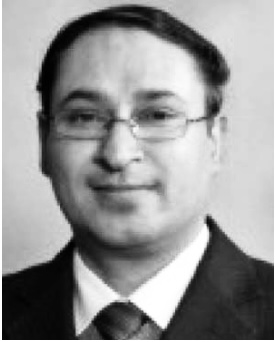}}]{Aneesh Krishna} is a Full Professor in the School of Electrical Engineering, Computing and Mathematical Sciences at Curtin University, Australia. He received his Ph.D. degree in Computer Science from the University of Wollongong, Australia. He has held several academic positions, including Lecturer in Software Engineering with the School of Computer Science and Software Engineering at the University of Wollongong from February 2006 to June 2009. His research interests include artificial intelligence for software engineering, model-driven development and evolution, requirements engineering, agent systems, formal methods, data mining, computer vision, machine learning, bioinformatics, and renewable energy systems. He has published more than 200 articles in reputed journals and international conference proceedings.
\end{IEEEbiography}
\vspace{-10mm}
\begin{IEEEbiography}
[{\includegraphics[width=1in,height=1.25in,clip,keepaspectratio]{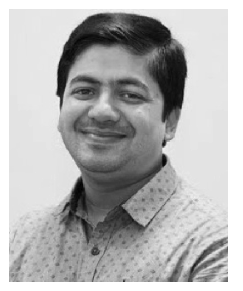}}]{Monowar Bhuyan} (WASP Fellow and Senior Member, IEEE) received his Ph.D. degree in Computer Science and Engineering from Tezpur University, Assam, India. He is currently an Associate Professor in the Department of Computing Science at Umeå University, Sweden. He established and leads the Cyber Analytics and Learning Group and is a Senior Member of the Autonomous Distributed Systems Lab. Prior to this, he held academic and research positions at several institutions, including the Nara Institute of Science and Technology, Japan; Assam Kaziranga University, India; and Umeå University, Sweden, at various levels from Junior Scientist to Associate Professor between January 2009 and December 2019. He has published more than 100 papers in leading peer-reviewed international journals and conference proceedings and authored the advanced textbook ``Network Traffic Anomaly Detection and Prevention'' with Springer. His experience in leading and co-leading research projects has attracted more than SEK 40 million in funding from national, European Commission, and international funding agencies. His research interests include machine learning, anomaly detection, systems and AI security, and distributed systems.
\end{IEEEbiography}
\vfill

\end{document}